\documentclass[11pt]{article}

\usepackage[final]{acl}

\usepackage{times}
\usepackage{latexsym}
\usepackage{amsmath}
\usepackage{amssymb}
\usepackage{amsfonts}
\usepackage{csquotes}
\usepackage{CJKutf8}

\usepackage{makecell}
\usepackage{longtable}
\usepackage{booktabs}
\usepackage{CJKutf8}
\usepackage{pdflscape}
\usepackage{caption}
\usepackage{multirow}
\usepackage{enumitem}
\usepackage{tabularx}  % For text wrapping
\usepackage{array}

\usepackage[T1]{fontenc}
\usepackage[utf8]{inputenc}

\usepackage{microtype}

\usepackage{inconsolata}

\usepackage{graphicx}

\usepackage{booktabs}
\usepackage[table]{xcolor}
\usepackage{cleveref}

\title{Latent Mechanisms of Language Control in Multilingual Language Models}
\author{Ryo Mitsuhashi\thanks{The first author was an intern at QCRI at the time this work was conducted.} \\
  Princeton University \\
  New Jersey, USA \\
  \texttt{rm4411@princeton.edu} \\
  \And
  Sabri Boughorbel \\
  Prince Sattam bin \\ Abdulaziz University \\
Riyadh, Saudi Arabia \\
\And
  Majd Hawasly \\
  QCRI, 
  Hamad Bin Khalifa \\University, 
  Doha, Qatar\\
  \texttt{mhawasly@hbku.edu.qa} }

\begin{document}
\maketitle
\begin{abstract}
  Multilingual large language models can exhibit \textit{unintended code-switching} -- unnecessarily alternating between languages during generation. We present a comparative study of three methods that identify language-controlling latents in cross-layer transcoders: activation value-based selection (\texttt{ValSel}), activation frequency-based selection (\texttt{FreqSel}), and LLM-generated latent annotation-based selection (\texttt{AnnSel}). To evaluate the efficacy of these methods in identifying language-controlling latents, we introduce two multilingual benchmarks that exhibit code-switching for fine-grained analysis of language steering across seven languages. Through targeted intervention experiments on \texttt{Gemma-2-2B} and \texttt{Qwen3-4B}, we find that all three methods effectively manipulate generation language, with \texttt{FreqSel} achieving the strongest overall performance, while \texttt{AnnSel} offering interpretable latent selection through explicit language annotations. A knock-out analysis suggests the methods select non-overlapping but each-functional latent subsets, indicating redundancy rather than a single canonical language direction. Code and data can be found at \url{https://github.com/rm-3284/Latent-Mechanism-Multilingual}.
\end{abstract}

\section{Introduction}
Large language models have achieved remarkable multilingual capabilities, but this flexibility comes with an unintended consequence: \textit{code-switching}, where models inappropriately alternate between languages mid-generation~\citep{marchisio2024language-confusion}. This problem motivated explicit language consistency rewards during training of models such as DeepSeek-R1~\citep{deepseekai2025deepseekr1incentivizingreasoningcapability}.

Recent mechanistic interpretability work has identified neurons and latents that causally influence output language~\citep{tang2024language-neurons, deng-etal-2025-unveiling}, suggesting that targeted interventions could provide fine-grained language control in generation. However, to our knowledge, no systematic comparison exists of language-controlling mechanism discovery, and it remains unclear whether different approaches uncover the same underlying latents or redundant representations.

We compare three approaches that identify language-specific latents in cross-layer transcoders (CLTs)~\citep{ameisen2025circuit}: \textbf{\texttt{ValSel}}, based on activation value differences~\citep{deng-etal-2025-unveiling}; \textbf{\texttt{FreqSel}}, based on language-specific activation frequency~\citep{andrylie2025sparseautoencoderscapturelanguagespecific}; and \textbf{\texttt{AnnSel}} that utilizes LLM-generated annotations to find latents whose descriptions explicitly reference the target language from an extracted circuit graph. Unlike prior work on single-layer sparse autoencoders (SAEs), CLTs provide a unified sparse dictionary across all layers, enabling holistic analysis of language representations in latent space.

The contributions of this work include:  a comparative study in controlled settings of language-steering latent selection methods; two controlled benchmarks; an analysis of intervention mechanism efficacy in 7 languages; and further, we highlight the potential redundancy of language representation in two families of models.

%The methods differ in how language-related mechanisms are identified, relying respectively on activation magnitudes, activation frequency, or latent annotation. We show that these approaches yield substantially different sets of latent variables, each capable of manipulating the target language, indicating a degree of latent redundancy. To facilitate a fine-grained study of code-switching, we introduce a dedicated dataset and evaluate the three approaches through targeted intervention experiments.

\section{Preliminaries}
\paragraph{Mechanistic Approaches to Language Control}

Prior work has investigated language control at multiple levels of abstraction. At the neuron level, \citet{tang2024language-neurons} identified language neurons whose selective activation steers output language. At the representation level, \citet{goncharov2025languagesteeringlatentspace} extracted language directions via PCA, enabling steering with directional ablation. More recently, sparse autoencoders (SAEs), which address \textit{polysemanticity} by learning an overcomplete sparse basis of monosemantic features~\citep{bricken2023monosemanticity}, have been applied to isolate language-controlling features through activation magnitude~\citep{deng-etal-2025-unveiling} or frequency patterns~\citep{andrylie2025sparseautoencoderscapturelanguagespecific}.

\paragraph{Cross-Layer Transcoders (CLTs)}
 extend SAEs by replacing MLP blocks across all layers with a jointly-trained sparse dictionary~\citep{ameisen2025circuit}, capturing features spanning model depth. In this paper, we use CLTs trained for \texttt{Gemma-2-2B}~\citep{lieberum-etal-2024-gemma} and \texttt{Qwen3-4B}~\cite{qwen3-4b-transcoders}.
\section{Manipulating Generation Language through Latent Interventions}
We study three language-specific \textbf{latent selection methods}, and four \textbf{latent intervention strategies} that steer generation toward a target language.
%We compare three latent selection methods that differ in how they characterize language specificity, followed by a unified intervention framework.
\subsection{Latent Selection Methods}
Given a CLT with latent space $\mathcal{S}$ and a collection of languages $L$, we identify for language $l \in L$ a subset of latents $\mathcal{S}^l \subset \mathcal{S}$ that are specifically associated with the generation of $l$. We compare three selection criteria based on activation magnitude, activation frequency, and semantic annotation.

\subsubsection{Value-Based Selection (\texttt{ValSel})}
%There has been recently a number of proposed approaches in the literature to language-controlling latent selection in SAEs. We present next a sample of reported methods and suggest a novel approach that leverage on latent annotations.  
%\subsubsection{Activation value-based latent selection}
% Inspired by the recent work of~\cite{deng-etal-2025-unveiling} and similar, for a latent $s$ and for all the tokens in a  dataset $D^l$ of a language  $l$, we compute the 
% %attribution graph and get the activation value of each latent. Then, calculate 
% the mean activation --- $\mu^l_s$ per latent for the language.

% Then,  for a collection of languages $L$, the $l$-monolinguality of a latent can be computed as $v_s^l = \mu^l_s - \frac{1}{|L|-1} \sum_{j \in L \setminus l} \mu^j_s$.

% The top-$K$ latents with the highest $l$-monolinguality $v^l_s$ are chosen per language $l$. We call this selection method \texttt{ValSel} in the rest of this paper.

Inspired by \citet{deng-etal-2025-unveiling}, this method identifies latents whose activation magnitude is distinctively higher for a target language compared to others. For a latent $s$ and a language-specific corpus $D^l$, we compute the mean activation:
\begin{equation}
    \mu^l_s = \frac{1}{|D^l|} \sum_{t \in D^l} a_s(t)
\end{equation}
where $a_s(t)$ denotes the activation of latent $s$ for token $t$. The \textit{l-monolinguality score} of latent $s$ for language $l$ is then defined as:
\begin{equation}
    v_s^l = \mu^l_s - \frac{1}{|L|-1} \sum_{j \in L \setminus \{l\}} \mu^j_s
\end{equation}
We select the top-$K$ latents with the highest $l$-monolinguality scores for each language $l$.

\subsubsection{Frequency-Based Selection (\texttt{FreqSel})}
% this is the other paper

% We call a latent $s$ active for a token $t$ if it has a non-zero activation value ( $s^t>0$). Inspired by the recent work of ~\cite{andrylie2025sparseautoencoderscapturelanguagespecific}, for a language $l$, we compute the \textit{activation probability} for $s$ in the language $l$ as $P^l_{s} = {|t \in D^l\;;\; s^t > 0|}/{ |D^l|}$.

% Then, for the language collection $L$, we compute a normalized activation probability vector for $s$,  $\mathbf{P}_s = (P^{l_1}_{s}, P^{l_2}_{s} , \ldots P^{|L|}_{s})$. We consider a latent to be \textit{specific} to a language $l$ if $P^{l}_{s} = \max \mathbf{P}_s$ and no other language $l'$ satisfies $P^{l'}_{s} \ge T \max {P^l_s}$ for some threshold $T\in[0,1]$. We select the top-$K$ specific latents to represent language $l$. We call this method \texttt{FreqSel} in the rest of this paper.

Following \citet{andrylie2025sparseautoencoderscapturelanguagespecific}, this method identifies latents that activate more frequently for a specific language. %A latent $s$ is considered \textit{active} for token $t$ if $a_s(t) > 0$.
We compute the \textit{activation probability} for latent $s$ in language $l$ as:
\begin{equation}
    P^l_s = \frac{1}{|D^l|}{|\{t \in D^l : a_s(t) > 0\}|}
\end{equation}
A latent is considered \textit{specific} to language $l$ if: (1) $P^l_s = \max_{j \in L} P^j_s$ and (2) no language $l' \neq l$ satisfies $P^{l'}_s \geq T \cdot P^l_s$ for a threshold $T \in [0,1]$. We select the language-specific latents that are active in at least $N\%$ of the examples and at least $M$\% of the tokens in the corpus of the language $l$.

% We consider a latent \textit{active} for a certain language $l$ if and only if it fires for more than $10\%$ of tokens  and at least $N\%$ of examples in $D^l$, the language-specific corpus.

\subsubsection{Annotation-Based Selection (\texttt{AnnSel})}

% Since we are interested in the features that control the output, we filter and preprocess the sentences so that the sentences are cut off in the middle and the next word that the model is supposed to predict is some word in the target language and not comma, period, or numbers.

We propose an approach that supports interpretability by leveraging LLM-generated latent annotations. This method operates in three stages: circuit tracing, path pruning, and annotation filtering.

For each sentence in $D^l$, we trace the attribution graph from input tokens to the predicted output token using Circuit Tracer~\citep{circuit-tracer}. We truncate sentences at a midpoint, ensuring the target token is a valid word in $l$. We then greedily prune the graph, retaining only its high-importance paths from each token embedding to the top predicted logit based on maximum bottleneck edge weight. Finally, we extract latents contributing to the prediction, filtering for those with annotations explicitly referencing language $l$ (e.g., containing ``Japanese'' or ``Japan'' for $l=$`\texttt{ja}') to ensure interpretable, language-related selection. Refer to Appendix~\ref{app:annsel_implementation} for more detail.

% For each sentence in a language-specific corpus $D^l$, we compute the latent attribution graph by tracing the circuit that generates the mid token given the first half of the sentence as a context. We prune the computed attribution graph by \textit{path importance}. More specifically, from each token embedding to the top logit of the predicted token, we greedily find distinct paths in the graph with maximum bottleneck edge weight, and set threshold to the minimum edge weight, the weights from embeddings to the first node, and from the last node to the final logit. After pruning non-important paths, we extract all the latents contributing to the last token position. 

% We aggregate all the latents found for all sentences in $D^l$ and extract their LLM annotations. For this, a number of highly-activating sentences per latent are extracted from a generic dataset, with the responsible tokens highlighted, and an LLM is tasked to describe the common pattern in the extracted examples. Finally, we filter out those latents whose annotations do not explicitly refer to the language $l$ to get the final latent selection per language. We call this method \texttt{AnnSel} in the rest of this paper.

\begin{table*}[ht!]

    \small
    \centering
    \begin{tabular}{l|c|c}
         \hline
         &Context& Target  \\ \hline
         \textbf{Antonyms} & \textsf{\small The opposite of }``\textbf{[adj]}'' \textsf{is ``} & \textbf{[antonym]} \\
            {\tiny\textcolor{gray}{ex1 (en, fr)}} & {\tiny\textcolor{gray}{\textit{\textsf{\footnotesize The opposite of ``\textbf{grand}'' is ``}}}} & \textcolor{gray}{\textit{\textbf{petit}}} \\
            
            {\tiny\textcolor{gray}{ex2 (de, ja)}}& {\tiny\textcolor{gray}{\textit{\textsf{\footnotesize Das Gegenteil von ``\textbf{\begin{CJK*}{UTF8}{goth}大きい\end{CJK*}}'' ist ``}}}} & \textcolor{gray}{\textit{{\begin{CJK*}{UTF8}{goth}\textbf{小さい}\end{CJK*}}}} \\ \hline
         
         \textbf{Enumerations} & \textsf{\small The [enum] are:} \textbf{[item~1], [item~2],}  & \textbf{[item~3], [item~4], \ldots} \\
              {\tiny\textcolor{gray}{ex3 (en, es)}}
              & {\tiny\textcolor{gray}{\textit{\textsf{\footnotesize The four seasons are:}}}} {\textcolor{gray}{\textit{\textbf{el verano}, \textbf{el otoño}, }}} & \textcolor{gray}{\textit{\textbf{el invierno, la primavera}}} \\
              
              {\tiny\textcolor{gray}{ex4 (fr, zh)}}& {\tiny\textcolor{gray}{\textit{\textsf{\footnotesize Les jours de la semaine sont: \textbf{\begin{CJK*}{UTF8}{goth}星期日\end{CJK*}}, \textbf{\begin{CJK*}{UTF8}{goth}星期一\end{CJK*}}, \textbf{\begin{CJK*}{UTF8}{goth}星期二\end{CJK*}}, }}}} & \textcolor{gray}{\textit{\textbf{\begin{CJK*}{UTF8}{goth}星期三\end{CJK*}},  \textbf{\begin{CJK*}{UTF8}{goth}星期四\end{CJK*}}, \ldots}} \\ \hline
    \end{tabular}
        \caption{Data format and examples of the two evaluation datasets. The sentence frame (in \textsf{\small sans-serif}) is in the context language $l_0$, while the target language $l$ is in \textbf{bold}. The examples show context/target languages ($l_0$, $l$).\label{tab:data_format}}

\end{table*}

\subsection{Intervention Strategies}
\label{sec:interventions}
%Using the extracted latents, we apply various types of interventions to encourage the generation of a language $l$ in a context of language $l_0$  as follows:
{For the sets of discovered language-specific latents $\{\mathcal{S}^l\}^{l\in L}$, we apply inference-time interventions to steer generation from a context language $l_0$ toward a target language $l$. We evaluate four  strategies:

\begin{itemize}[nosep,leftmargin=*]
% \itemsep0em
% \topsep0em
% \parsep0em
% \partopsep0em
\item{\textbf{Target Amplification}:} Set each latent $s \in \mathcal{S}^l$ to its mean nonzero activation computed over $D^l$. This naturally amplifies the signal for the target language $l$  while avoiding out-of-distribution effects, without affecting other languages.
    \item{\textbf{Distractor Zero Ablation}:} Suppress competing signals from other languages by zeroing all latents associated with the distractor $l_0$, or the latents of all non-target languages $\bigcup_{j \in L \setminus \{l\}} \mathcal{S}^j$.
    % Set to zero all latents associated with the distractor language $l_0$, or the latents of all non-target languages $\bigcup_{j \in L \setminus \{l\}} \mathcal{S}^j$. This suppresses competing signals from other languages.
    \item{\textbf{Distractor Direction Ablation}:} Ablate the latents of the distractor language $\mathcal{S}^{l_0}$ by \textit{direction ablation}, where the context language gets suppressed by removing the projection of the residual stream activation along a latent's direction. We do this either for the single layer with the most latents (\textit{one-layer}), or for all layers (\textit{multi-layer}).
    \item{\textbf{Combined Interventions}:} Apply distractor ablation and target amplification simultaneously, boosting the signal of the target language while suppressing that of the context language. We test distractor zero ablation + target amplification (\textbf{Zero+Amp}), and distractor one-layer direction ablation + target amplification (\textbf{1L+Amp}).
    %Simultaneously apply a form of distractor ablation along with target amplification,  boosting the signal of the target language while suppressing that of the context language. We test distractor zero ablation + target amplification (\textbf{Zero+Amp}), and distractor one-layer direction ablation + target amplification (\textbf{1L+Amp}).
\end{itemize}

% \begin{description}
    % \paragraph{Target Amplification} Set each latent $s \in \mathcal{S}^l$ to its mean nonzero activation computed over $D^l$. This naturally amplifies the signal for the target language $l$  while avoiding out-of-distribution effects, without affecting other languages.
    % \paragraph{Zero Ablation} Set to zero all latents associated with the distractor language $l_0$, or the latents of all non-target languages $\bigcup_{j \in L \setminus \{l\}} \mathcal{S}^j$. This suppresses competing signals from other languages.
    % \paragraph{Distractor Direction Ablation} Ablate only the latents of the distractor language $\mathcal{S}^{l_0}$ by \textit{direction ablation}, where the projection of the residual stream activation along a latent's direction is fully removed, completely suppressing the context language. We do this either for the latents of the layer with most latents (single-layer), or for all-layers.
    % \paragraph{Combined Interventions} Simultaneously apply a form of distractor ablation with target amplification,  boosting the signal of the target language  while suppressing the context language's. We test zero distractor ablation + target amplification (\textbf{Zero+Amp}), and  distractor one-layer direction ablation + target amplification (\textbf{1L+Amp}).
% \end{description}
}

% \begin{itemize}
%     \item \textbf{Amplification}: setting the extracted latents of the target language $l$ to their mean nonzero activation value in all the tokens of language $l$.
%     \item \textbf{Ablation}: ablating the latents of all other languages except the target language, $L\setminus l$.
%     \item \textbf{Distractor ablation}: ablation of the latents of the context language  $l_0$ only.   
%     \item \textbf{Amplification + distractor Ablation}: at the same time, ablation of the latents of the context language $l_0$  and setting the latents of the target language $l$ to the mean activation of these latents for all the tokens of language $l$.
% \end{itemize}

\section{Experiments}

% \subsection{Experimental Setup}
All the experiments are conducted on the transcoders of \texttt{Gemma-2-2B-pt}\footnote{https://hf.co/mwhanna/gemma-scope-transcoders} from Gemma-scope~\cite{lieberum-etal-2024-gemma} and \texttt{Qwen3-4b}\footnote{https://hf.co/mwhanna/qwen3-4b-transcoders}.

For \texttt{ValSel}, we use $K=50$ (to parallel the 2 latents per layer choice in \citet{deng-etal-2025-unveiling} for the 26 layers of \texttt{Gemma-2-2B}). For \texttt{FreqSel}, we set $T = 0.8$, $N = 98$, and $M = 10$, following \citet{andrylie2025sparseautoencoderscapturelanguagespecific}. For \texttt{AnnSel}, we utilize the Circuit Tracer tool~\cite{circuit-tracer} and  Neuronpedia annotations\footnote{https://www.neuronpedia.org}. 

To extract the latent sets, we use FLORES+~\cite{nllb-24} in seven languages: 
% used for evaluation
 English (en), French (fr), German (de), Spanish (es), Chinese (zh), Korean (ko), and Japanese (ja). We apply the three selection methods separately to both CLTs and each language subset.
We report in the Appendix (\Cref{tab:latent_counts})  the count of latents extracted per method and language. We also report the computational cost of each method in Appendix~\ref{app:cost} and hyperparameter sensitivity analysis in Appendix~\ref{app:hyperparam_sensitivity}. %Note that \texttt{ValSel} extracts 50 latents by design.

\subsection{Datasets}

We introduce \textbf{Antonyms} and \textbf{Enumerations}, two novel evaluation datasets for language-steering interventions.\footnote{Refer to Appendix~\ref{app:antonyms} and \ref{app:enumerations} for the complete list. The datesets can also be accessed from the project's repository.} The \textbf{Antonyms} task requires producing a target antonym in language $l$ given a context in language $l_0$, creating a controlled code-switching scenario. \textbf{Enumerations} tests open-ended generation by requiring the continuation of a list in the target language. Table~\ref{tab:data_format} displays the data formats and examples.

%We introduce \textbf{Antonyms} and \textbf{Enumerations}, two novel evaluation datasets designed to measure the effectiveness of language-steering interventions.\footnote{The datasets will be publicly released.} \textbf{Antonyms} task requires producing a target antonym in a desired language $l$ given a sentence context in a different language $l_0$, creating a controlled code-switching scenario, while \textbf{Enumerations} asks for a continuation of a listing in the desired language, testing open-ended generation. The data format and some examples of the two datasets are shown in Table~\ref{tab:data_format}.

\paragraph{Dataset construction.} We collect 100 antonym pairs per language for the seven languages (en, fr, de, es, zh, ja, ko) for \textbf{Antonyms} and 7 ordered categories for \textbf{Enumerations}. The complete lists of adjectives and words are provided in Appendix~\ref{app:antonyms} and \ref{app:enumerations} respectively. For each of the $7 \times 6 = 42$ ordered pairs of context and target languages $(l_0, l)$ where $l_0 \neq l$, we generate a test example per antonym pair and per enumeration category.

\paragraph{Evaluation protocol.} Given a context, we evaluate whether the model prefers the target antonym or enumeration in language $l$ over its translations. We record the target token's \textit{logit} for \textbf{Antonyms} and length-normalized \textit{logprob} for \textbf{Enumerations} for each example across all seven languages. A successful intervention must increase these values relative to the baseline.

%Given a context, we measure whether the model assigns a higher probability to the target antonym or enumeration in language $l$ compared to its translations in other languages. For each example, we record the \textit{logit} assigned to the target token for \textbf{Antonyms} and the length-normalized \textit{logprob} for \textbf{Enumerations} across all seven languages. A successful intervention should increase the target logit or logprob post-intervention relative to the baseline before the intervention.

\begin{table*}[!htp]\centering
\resizebox{.96\textwidth}{!}{
\begin{tabular}{l|ccccccccccc}\toprule
&Distractor ($l_0$)&$L \setminus \{l\}$  &Distractor ($l_0$)  &Distractor ($l_0$)  &Target ($l$) &Zero+Amp &1L+Amp \\
& zero  & zero  & one-layer  & multi-layer    & amplification & &  \\

&ablation &ablation & direction ablation & direction ablation  & &&

\\\midrule
% \texttt{ValSel} &16.42 &22.56 &5.01 &46.41  &69.10 &\textbf{79.83} &\underline{69.19} \\

\texttt{ValSel} &-8.02 &22.39 &1.99 &13.98 &69.57 &\textbf{79.83} &\underline{69.73}  \\

%-8.06 &22.38 &1.98 &-6.11 &13.96 &-37.09 &-64.91 &69.57 &\textbf{79.83} &\underline{69.74} \\
% \texttt{FreqSel} &26.12 &22.10 &1.58 &-22.06 &93.79 &\textbf{99.04} &\underline{96.32}\\

\texttt{FreqSel} &2.02 &21.39 &-2.95 &-51.62 &90.83 &\textbf{99.07} &\underline{96.16} \\

% 2.00 &21.37 &-2.98 &-38.37 &-51.64 &-134.69 &-218.83 &90.83 &\textbf{99.04} &\underline{96.12} \\
% \texttt{AnnSel} & 16.46 &13.04 &0.81 &70.81 &85.05 &\textbf{96.69} &\underline{86.77} \\

\texttt{AnnSel}  &-0.20 &11.90 &0.30 &64.05 &87.00 &\textbf{96.72 }&\underline{89.33} \\

% -0.22 & 11.9 & 0.28 & -4.79 & 64.03 & -38.16 & -180.44 & 87.01 & \textbf{96.69} & \underline{89.31} \\
\bottomrule
\end{tabular}
}
\caption{Intervention effectiveness per method using \textbf{Antonyms}, computed as the total change in top-1 logit margin under the intervention (Equation~\ref{eq:change}) averaged over all target languages (bigger is better) for \texttt{Gemma-2-2B}.\label{tab:interventions}}
\end{table*}

\subsection{Findings}

\subsubsection{Intervention strategy effectiveness}
To assess language manipulation efficacy, we intervene on the extracted latents, enforcing a target language $l$ in a sentence with context language $l_0$ using \textbf{Antonyms}, where the target is a single token. We measure the target token's \textbf{\textit{top-1 logit margin}}, which is the separation of the logit of the target adjective from its best competitor: $
m_l = \max_{i \in L \setminus \{l\}} a_i - a_l  
$, 
where $a_l$ is the logit of the target adjective in language $l$. A more negative $m_l$ value indicates a higher probability for target language $l$. We measure the change in $m_l$ due to the intervention: \begin{equation}
    \Delta m_l = - (m_l^{after} - m_l^{before})
    \label{eq:change}
\end{equation}
Hence, a larger positive change indicates a more effective intervention. 
Summary results for \texttt{Gemma-2-2B} are in \Cref{tab:interventions}, with detailed results in the Appendix  \Cref{tab:int_V,tab:int_F,tab:int_A}, as well as results for \texttt{Qwen3-4B} (\Cref{tab:intQ_A,tab:intQ_F,tab:intQ_V}). 

The most effective intervention is \textbf{Zero+Amp}, combining zero ablation of distractor language $l_0$  with target language amplification, followed by \textbf{1L+Amp}, which instead uses single-layer directional ablation of $l_0$. Zero ablation seems to outperform the more powerful direction ablation, possibly due to the latent overlap of related languages, which weakens the target language when direction-ablating the context language.

\subsubsection{Selection method efficacy}
We now apply the most effective intervention (\textbf{Zero+Amp}) to all selection methods and languages to contrast their performance per target language. Table~\ref{tab:total_logits} shows summary results for \texttt{Gemma-2-2B} on \textbf{Antonyms} and \textbf{Enumerations} (detailed results in~\Cref{tab:valsel_logits,tab:freqsel_logits,tab:annsel_logits} in the Appendix), and results for \texttt{Qwen3-4B} are in \Cref{tab:total_logits_qwen}.

\begin{table}[ht!]
    % \centering

    \begin{center}\resizebox{0.48\textwidth}{!}{\begin{tabular}{r|r|ccccccc|c}
    \toprule
        &&es&en&zh&de& ja&fr&ko & Avg\\
        \hline
\multirow{3}{.4cm}{\rotatebox[origin=c]{90}{\textbf{Antons.}}}
&\texttt{ValSel} &\textbf{14.19} &\textbf{10.16} &16.06 &\underline{13.56} &4.63 &\underline{9.01} &\underline{12.18} &11.40 \\
&\texttt{FreqSel} &8.95 &4.34 &\textbf{34.50} &7.85 &\underline{8.65} &6.45 &\textbf{28.30 }&\textbf{14.15} \\
&\texttt{AnnSel} &\underline{13.86} &\underline{4.87} &\underline{24.52} &\textbf{19.67} &\textbf{11.12} &\textbf{13.54} &9.11 &\underline{13.81} \\
\bottomrule    
    % \end{tabular}}\end{center}  
    % \begin{center}
    % \resizebox{0.47\textwidth}{!}{\begin{tabular}{r|ccccccc|c}
    \toprule
        % &es&en&zh&de& ja&fr&ko & Avg\\
        % \hline
\multirow{3}{.4cm}{\rotatebox[origin=c]{90}{\textbf{Enums.}}}&\texttt{ValSel} &\underline{1.00} &\underline{-0.08} &\underline{1.13} &0.93 &\underline{1.54} &0.72 &\underline{1.87} &\underline{1.02} \\
&\texttt{FreqSel} &\textbf{2.08} &-0.33 &\textbf{2.12} &\textbf{2.69} &\textbf{2.52 }&\textbf{1.40} &\textbf{3.92} &\textbf{2.06} \\
&\texttt{AnnSel} &0.90 &\textbf{0.30} &1.05 &\underline{1.83} &1.47 &\underline{0.85} &0.38 &0.99 \\
\bottomrule        
    \end{tabular}}\end{center}
         \caption{Selection method efficacy per language under the \textbf{Zero+Amp} intervention for \texttt{Gemma-2-2B}. \textit{Top}: post-intervention total change in top-1 logit margin for \textbf{Antonyms}; \textit{Bottom}: total change in normalized logprob of the target sequence for \textbf{Enumerations}.}

    % (Equation~\ref{eq:change}).}
    \label{tab:total_logits}
\end{table}

In total across all settings and models, all the selection methods are effective. \texttt{FreqSel} achieves the greatest total absolute logit and logprob change across the tasks, with significant performance gains in Asian languages (especially Chinese and Korean). \texttt{AnnSel}, on the other hand, shows decent performance across all languages but is hindered by its weak performance in Korean, which might be justified by the small number of extracted latents for Korean (\Cref{tab:latent_counts}). We also present the effectiveness of the latents found in a free generation setting in Appendix~\ref{app:cont_generation}.

\subsubsection{Is there redundancy in latent language representation?}
To evaluate the equivalence of latent sets from different selection methods, we conduct a cross-method knock-out experiment for each target language. We apply the amplification intervention using the first method's latents, then ablate the residual stream change vector's projection onto the second method's representative direction. This isolates whether the first method's latents can still drive a positive logit margin change independent of the second method.

\begin{table}[!htp]\centering\small
\label{tab:knonkout}
 \resizebox{0.41\textwidth}{!}
{ 
\begin{tabular}{c|llrr}\toprule
&Amplification  &Ablation  &Total   \\
& method &  method & $\Delta m_l$ \\
\midrule
\multirow{3}{.4cm}{\rotatebox[origin=c]{90}{\textbf{Gemma-2-2B}}}&AnnSel &FreqSel &5.20 \\
&AnnSel &ValSel &38.82 \\
&FreqSel &AnnSel &\underline{90.63} \\
&FreqSel &ValSel &\textbf{92.16} \\
&ValSel &AnnSel &49.52 \\
&ValSel &FreqSel &-1.19 \\
\midrule 
\multirow{3}{.4cm}{\rotatebox[origin=c]{90}{\textbf{Qwen3-4B~}}}&AnnSel &FreqSel &1.28 \\
&AnnSel &ValSel &7.58 \\
&FreqSel &AnnSel &\textbf{155.29} \\
&FreqSel &ValSel &85.82 \\
&ValSel &AnnSel &\underline{107.55} \\
&ValSel &FreqSel &33.86 \\
\bottomrule
\end{tabular}
}
\caption{Aggregate knock-out experiment results over all languages for \texttt{Gemma-2-2B} and \texttt{Qwen3-4B}, where the latents of one selection method for the target language are amplified and the latents of the second method are used for direction ablation.}
\end{table}

We report in \Cref{tab:knonkout} the total logit margin change aggregated over the seven languages per method pair for \texttt{Gemma-2-2B} and \texttt{Qwen3-4B} on \textbf{Antonyms}. The generally positive numbers indicate a form of robustness, however, the asymmetry of the results for the same pair of methods requires further investigation, as it indicates latent sets that are not completely independent. We provide 
in the Appendix~\ref{apx:redundancy} additional observations related to the cosine similarity of the latent set representative residual-stream directions where language directions vary to different extents in their similarity,  in addition to observations related to latent annotations where the intersection of latent sets of the same language identified by different methods is not empty.

\section{Conclusion}

We presented a comparative study of three methods for identifying language-specific latents in cross-layer transcoders: value-based  (\texttt{ValSel}), frequency-based  (\texttt{FreqSel}), and annotation-based  (\texttt{AnnSel}). Through experiments on the controlled \textbf{Antonyms} and \textbf{Enumerations} tasks in seven languages, we demonstrated that all three methods can effectively steer generation language. \texttt{FreqSel} achieves the strongest overall performance, particularly for Asian languages, while \texttt{AnnSel} provides the added benefit of interpretable latent selection.

Future work could look into generalizing the findings of this work to larger models and to more languages. Real-world code-switching scenarios may present different challenges to the studied controlled settings. A deeper understanding of to what extent different methods identify similar or different latents has implications for both understanding multilingual representations and developing robust language-steering interventions.

\section{Limitations}
\label{sec:limitations}

Our study has several limitations. First, experiments were conducted on the CLTs of two small models (\texttt{Gemma-2-2B} and \texttt{Qwen3-4B}); findings may not generalize to other architectures or standard single-layer SAEs. Second, the evaluation was limited to seven languages, predominantly high-resource, and two controlled code-switching tasks (single-token and multi-token), although we also present an experiment in a less constrained setting in Appendix~\ref{app:cont_generation}. Real-world code-switching scenarios may present different challenges. Third, \texttt{AnnSel}'s effectiveness depends on the quality of Neuronpedia annotations, which may be inconsistent across languages, as evidenced by the low latent count for Korean (7 vs. 45+ for other languages). Also, keyword matching may not be the most effective selection method, as it does not consider annotation context (e.g., it selects "Non-English characters" as an English latent). Finally, we did not systematically evaluate computational costs, though \texttt{AnnSel} requires additional circuit tracing compared to the statistics-based methods. We do present a comparison of computational costs in Appendix~\ref{app:cost}.

\section*{Potential Risks}
 
Language-steering interventions could in principle be repurposed to force generation in a specific language for deceptive purposes or to suppress a language entirely. We consider this risk to be low in practice, as the interventions require direct access to model internals (CLT weights and latent activations) and are demonstrated only on small-scale models in controlled settings. The benchmarks and methods presented in this work are intended to advance mechanistic understanding of multilingual representations and do not introduce new capabilities beyond what is already achievable through prompting or fine-tuning.
\bibliography{custom}

\newpage
\appendix
\onecolumn

\section{\texttt{AnnSel} implementation details}
\label{app:annsel_implementation}

The implementation of \texttt{AnnSel} can be decomposed into three steps: attribution graph construction, path pruning, and annotation-based latent selection.

\subsection{Attribution graph construction}

Since attribution graphs explain how the model predicts the next token, given a sentence from FLORES+~\citep{nllb-24}, we randomly cut the sentence in the middle so that the next token the model should predict is a valid word in the target language, i.e., not a number or punctuation.

We use implementation from Circuit Tracer~\cite{circuit-tracer} with parameters: $\texttt{max\_n\_logits} = 5, \texttt{desired\_logit\_prob} = 0.95, \texttt{max\_feature\_nodes} = \mathrm{None}$. It means the attribution graphs are computed to explain the top $n$ logits, where $n = \min (5, m)$ and $m$ is the smallest integer that satisfies $\sum_{i=1}^m prob_i \ge 0.95$. We then prune the graph with $\texttt{node\_threshold} = 0.8, \texttt{edge\_threshold} = 0.98$, meaning keeping nodes that explain 80\% of total influence on the output logits and edges that explain 98\% of total influence on the output logits.

\subsection{Path Pruning}

After computing the attribution graph, we prune the graphs to prevent unimportant features from being included in the language supernodes. From each token to the top logit, we select important paths as follows. We first order the edges by the edge weight in descending order and keep choosing edges that are incident to already chosen nodes, starting from the token. When we reach the top logit, we can form a path from token to logit. Then, we remove that chosen path and start again. We keep doing this until either the edge weight chosen is less than or equal to 0.1 or we choose 75 distinct paths. We do this for every token position and obtain a set of paths.

After obtaining important paths, we further choose more important paths by setting a threshold on the weights of the first edge and last edge, since they represent the importance of the chosen paths in terms of the token and logit, respectively, and thus are directly involved in the information flow from token to logit. We choose paths whose first edge has more than 0.5 edge weight and whose last edge has more than 0.25 edge weight. Then, out of the chosen features, we extract all the features in the last token position and count how many times each of them appeared.

\subsection{Annotation-based latent selection}

After obtaining the set of features and the number of times they appear in the important paths, we first extract the description of each feature from Neuronpedia. We then remove all the features that does not mention the language name or highly relevant country name, e.g., ["Japanese", "japanese", "Japan", "japan"] for Japanese.

Then, what we have is a list of features that includes language names or country names in its description and their frequency. Let us denote the list by $\ell$ and the frequency of feature $f$ by $freq_f$. We then choose $f$ that satisfies $freq_f \ge 0.1 \max_{k \in \ell} freq_k$.

\section{Annotation Quality of \texttt{AnnSel}}

As we mention in Limitations (\Cref{sec:limitations}), the effectiveness of latents found by \texttt{AnnSel} depends on the quality of Neuronpedia annotations, and we consider this to be the primary reason why Korean latents found by \texttt{AnnSel} do not perform well. Also, the \texttt{AnnSel} criterion is quite simple, as we utilize keyword matching. In this paper, we present \texttt{AnnSel} as an interpretable alternative to other methods. However, more comprehensive validation would be necessary for other applications, as the quality and accuracy of latent annotations can substantially affect the coverage and reliability of the extracted latent sets.

We list below some annotations from Neuronpedia of Gemma-2-2B CLT latents that were selected by \texttt{AnnSel} to show that our simple criterion manages to capture semantically meaningful latents:
\begin{itemize}[nosep,leftmargin=*]
    \item German: “sentences in German that uses modal verbs to describe realistic situations”
    \item English: “words that are common in conversational English grammar despite having little content.”
    \item Spanish: “news casts, TV programs, and names of Spanish-speaking people in Panama”
    \item French: “French language and words related to treaties.”
    \item Japanese: “Japanese text related to programming errors, particularly requests for help”
    \item Korean: “Korean names, especially those with multiple parts separated by hyphens, in credits or medical contexts.”
    \item Chinese: “text and code snippets that occur in Chinese contexts”
\end{itemize}

We also observe that keyword matching may not be semantically correct. For example, the algorithm selects a latent with the description "Non-English characters" as an English latent since it matches the keyword "English". We do not remove these latents to keep the algorithm scalable.

\section{Applying SAE feature selection methods on CLT}
We apply two methods, \texttt{ValSel} and \texttt{FreqSel}, from \citet{deng-etal-2025-unveiling} and \citet{andrylie2025sparseautoencoderscapturelanguagespecific} respectively on the CLT. Since the original paper executes these methods on SAE, in order to justify the effectiveness of the methods on CLT, we conduct two experiments proposed in \citet{deng-etal-2025-unveiling} and \citet{andrylie2025sparseautoencoderscapturelanguagespecific}.

\subsection{\texttt{ValSel} and code-switching experiment}

\begin{figure*}
    \centering
    \includegraphics[width=\textwidth]{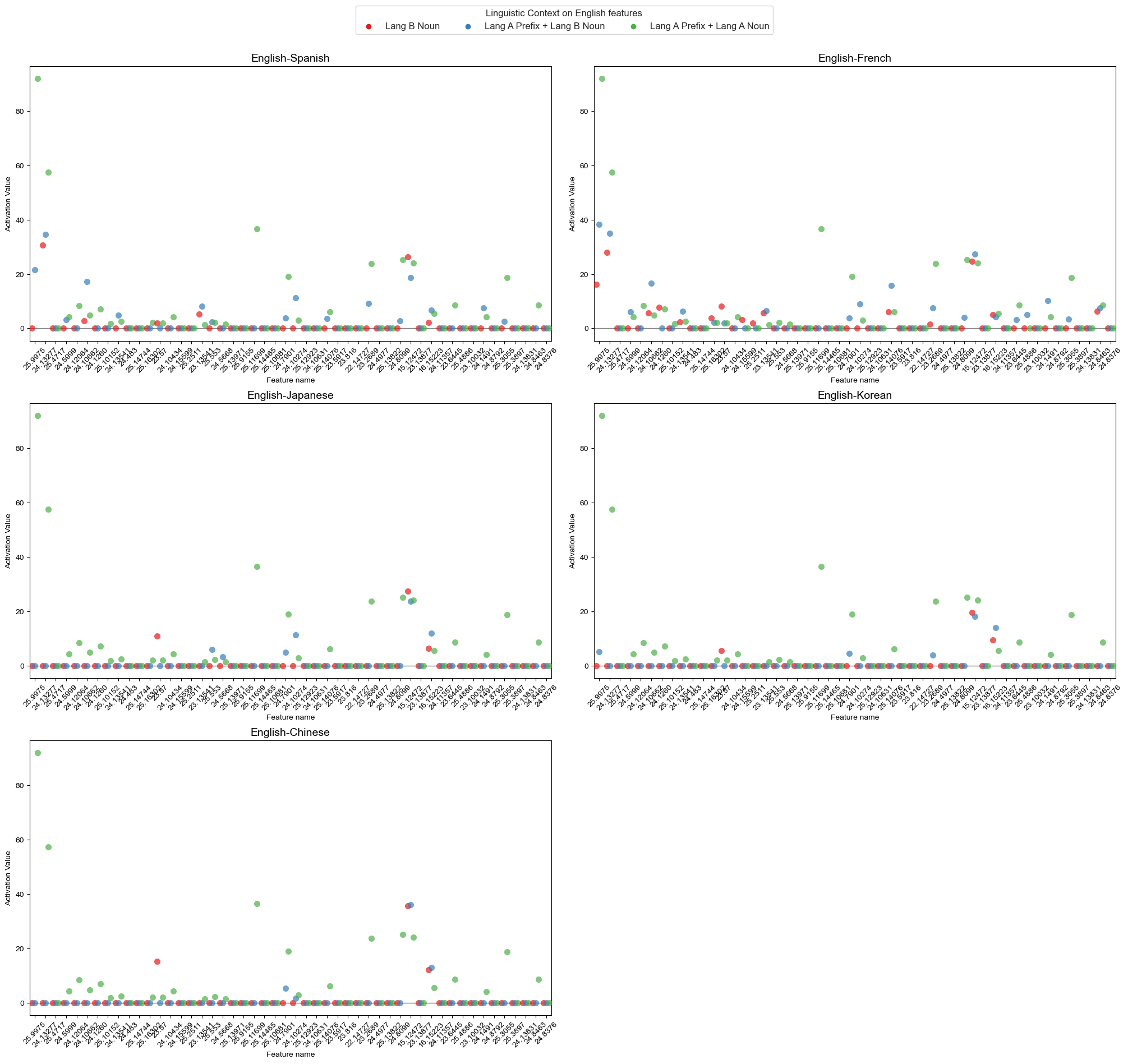}
    \caption{Activation values of English features on other language contexts and nouns of \texttt{Gemma-2-2B}.}
    \label{fig:code_switch_en}
\end{figure*}

\citet{deng-etal-2025-unveiling} confirms the specificity of the language-specific features by comparing the activation values in the normal context and code-switching context. Following their experiments, we use their code-switching dataset and plot the activation values of the top-50 features found and used for steering in our experiments. Keep in mind that we plot the activations for all the languages used in our experiment except German because \citet{deng-etal-2025-unveiling} does not include German in the experiments. Also, since we conduct the experiments on the CLT, the extracted features are not distributed uniformly across layers. The trend is the same as what is claimed in \citet{deng-etal-2025-unveiling}, and the activation is generally highest in the original context and noun (Lang A Prefix + Lang A Noun), somewhat active in original context and modified noun (Lang A Prefix + Lang B Noun), and close to zero in the modified noun (Lang B Noun). The results for English as Lang A are shown in \Cref{fig:code_switch_en}. For results in other languages, please refer to our codebase.

% \begin{figure*}
%     \centering
%     \includegraphics[width=\textwidth]{figures/code_switch_activations_fr.png}
%     \caption{Activation values of French features on other language contexts and nouns of \texttt{Gemma-2-2B}.}
%     \label{fig:code_switch_fr}
% \end{figure*}

% \begin{figure*}
%     \centering
%     \includegraphics[width=\textwidth]{figures/code_switch_activations_es.png}
%     \caption{Activation values of Spanish features on other language contexts and nouns of \texttt{Gemma-2-2B}.}
%     \label{fig:code_switch_es}
% \end{figure*}

% \begin{figure*}
%     \centering
%     \includegraphics[width=\textwidth]{figures/code_switch_activations_zh.png}
%     \caption{Activation values of Chinese features on other language contexts and nouns of \texttt{Gemma-2-2B}.}
%     \label{fig:code_switch_zh}
% \end{figure*}

% \begin{figure*}
%     \centering
%     \includegraphics[width=\textwidth]{figures/code_switch_activations_ja.png}
%     \caption{Activation values of Japanese features on other language contexts and nouns of \texttt{Gemma-2-2B}.}
%     \label{fig:code_switch_ja}
% \end{figure*}

% \begin{figure*}
%     \centering
%     \includegraphics[width=\textwidth]{figures/code_switch_activations_ko.png}
%     \caption{Activation values of Korean features on other language contexts and nouns of \texttt{Gemma-2-2B}.}
%     \label{fig:code_switch_ko}
% \end{figure*}

\subsection{\texttt{FreqSel} and text-generation experiment}

\citet{andrylie2025sparseautoencoderscapturelanguagespecific} confirms the downstream effectiveness of the features by steering and doing unconditional text-generation. Their steering is $\boldsymbol{x}_{new} = \boldsymbol{x} + \alpha \boldsymbol{z}_j^{max} \boldsymbol{d}^j$, where $\alpha \in \mathbb{R}$ is a scaling factor and $\boldsymbol{z}_j^{max}$ is the maximum value of the feature of interest $\boldsymbol{d}^j$ in the multilingual corpora. We use the same multilingual corpora used to determine the language-specific features when deciding on $\boldsymbol{z}_j^{max}$. The model was prompted with [BOS] and decoded with top-p sampling. Some examples are shown in \Cref{tab:text-generation}. The output is steered toward the target languages, although the sentences might not be perfectly natural.

\begin{table*}[h]
    \caption{Examples of unconditional text generation given [BOS] ($p=0.0$ means greedy decoding) of \texttt{Gemma-2-2B}}
    \centering\small
    \begin{tabular}{c|c|c|c}
        \hline
        \makecell{Intervened\\Language} & $\alpha$ & p & Output \\
        \hline
        de & 0.8 & 0.9 & \makecell{package Produktion Kleidung refroidissement nôtre enfans \\hindurchholen nôtre bissunter Bedürf Flüssigkeit geben Flüssigkeit Kleidung} \\
        \hline
        en & 0.1 & 0.0 & How to get the value of a variable in a function in python\textbackslash n\textbackslash n \\
        \hline
        es & 0.8 & 0.9 & \makecell{How convert Activity anuales in española y y argentina\textbackslash n \\pre acuerdos efectivos} \\
        \hline
        fr & 0.4 & 0.8 & \makecell{Que efficaces financières russes russes efficaces financières \\financières efficaces financières efficaces financières efficaces} \\
        \hline
        ja & 0.3 & 0.0 & How toget\begin{CJK}{UTF8}{min}指定したフォルダ内のファイル一覧をスクスクスクスク\end{CJK} \\
        \hline
        ko & 0.5 & 0.0 & How to 3D \begin{CJK}{UTF8}{mj}렌샷 것이다 렌샷 것이다 렌\end{CJK} \\
        \hline
        zh & 0.5 & 0.9 & The\begin{CJK}{UTF8}{min}特平阶在道路上有什么用呢？为什么不把柏的吗\end{CJK} \\
        \hline
    \end{tabular}
    \label{tab:text-generation}
\end{table*}

\section{Latent counts}
\Cref{tab:latent_counts} shows the number of extracted token per method and language pair for \texttt{Gemma-2-2B} and \texttt{Qwen3-4B}. Note that \texttt{ValSel} extracts 50 latents by design, while the other two methods return a variable number of latents that relates to the CLT training data.

\begin{table*}[ht]
    \caption{Count of extracted latents per method and language for \texttt{Gemma-2-2B} and \texttt{Qwen3-4B}.}
\small
    \centering
    \begin{tabular}{r|r|ccccccc}
    \toprule
        &&en&fr&de&es& zh&ko&ja \\
        \hline
     \multirow{3}{.3cm}{\rotatebox[origin=c]{90}{\texttt{Gemma}}}  &\texttt{ValSel}  & 50 & 50 & 50& 50 & 50 &50 & 50  \\
       & \texttt{FreqSel} & 101 & 423 & 579 & 509 & 268 & 938 & 234 \\
       & \texttt{AnnSel} & 17 & 41 & 59 & 19 & 41 & 7 & 45\\ \midrule

    \multirow{3}{.3cm}{\rotatebox[origin=c]{90}{\texttt{Qwen}}}  &\texttt{ValSel}  & 50 & 50 & 50& 50 & 50 &50 & 50  \\
       & \texttt{FreqSel} & 11 & 438& 500 & 299& 136& 366 & 293 \\
       & \texttt{AnnSel} & 12 & 16 &13&10&16&8&13\\ \midrule
       
    \end{tabular}

    \label{tab:latent_counts}
\end{table*}

\section{Computational Cost of Each Method}
\label{app:cost}

We report the computational cost in \Cref{tab:cost_comparison}. The numbers reported are cumulative costs in GPU-hours and kWh when processing 100 sentences $\times$ 7 languages on an NVIDIA L40 GPU. We use $\texttt{kWh} \approx \texttt{GPU-hours} \times (\texttt{util}/100) \times 0.3 \texttt{kW}$ to estimate kWh, as the TDP of the L40 is about 300 W. Since power does not scale linearly with utilization, this is likely a lower bound of the actual energy consumption.

\begin{table}[h]
\caption{Comparison of energy spent for each method and each model in GPU-hours and kWh}
\centering
\small
\begin{tabular}{lcc}
 & Gemma-2-2B & Qwen3-4B \\
\hline
\texttt{AnnSel} & 54 GPU-h, 4.4 kWh & 107 GPU-h, 11.8 kWh \\
\texttt{ValSel} & 5 GPU-h, 0.8 kWh & 19 GPU-h, 3.8 kWh \\
\texttt{FreqSel} & 5 GPU-h, 0.8 kWh & 23 GPU-h, 4.2 kWh \\
\end{tabular}
\label{tab:cost_comparison}
\end{table}

\noindent While the cost of \texttt{ValSel} and \texttt{FreqSel} mainly comes from the forward pass through the model and CLT, \texttt{AnnSel} requires more extensive circuit tracing and graph pruning on top of that. The additional cost of Qwen comes from the larger number of features in CLT. Specifically, the CLT for Qwen has 163,840 features per layer and 36 layers, while the CLT for Gemma has 16,384 features per layer and 26 layers in total.

\section{Hyperparameter Sensitivity Analysis}
\label{app:hyperparam_sensitivity}

We conduct the hyperparameter sensitivity analysis using Gemma-2-2B and the Antonyms dataset. All the reported numbers are the results of the Zero+Amp intervention.

\subsection{Fixed budget comparison}

As the number of latents found by each method differs, we fixed the latent budget per language to the smallest number identified across the three methods. \Cref{tab:matched_budget} shows the results. As \Cref{tab:latent_counts} shows, \texttt{AnnSel} has the fewest latent counts in all the languages except German. As for German latents, we rank the \texttt{AnnSel} latents by frequency and take the top 50 latents to match the size of \texttt{ValSel} latents.

\begin{table}[h]
\caption{The average total change per language for the three selection methods for Gemma-2-2B and the Antonyms dataset.}
\centering
\small
\begin{tabular}{lcccccccc}
 & de & en & es & fr & ja & ko & zh & Avg \\
\hline
\texttt{AnnSel} & \textbf{19.45} & 4.87 & 13.48 & \textbf{13.36} & \textbf{11.21} & 9.05 & \textbf{24.68} & \textbf{13.73} \\
\texttt{FreqSel} & 10.67 & 4.14 & 12.70 & 11.03 & 4.06 & \textbf{12.10} & 19.32 & 10.57 \\
\texttt{ValSel} & 13.63 & \textbf{12.92} & \textbf{14.22} & 9.37 & 4.04 & 9.06 & 17.27 & 11.50 \\
\hline
\end{tabular}
\label{tab:matched_budget}
\end{table}

\subsection{\texttt{ValSel} sensivity analysis}

The only hyperparameter used for \texttt{ValSel} is $k$ when choosing the top-$k$ highest latents. \Cref{tab:valsel-top-k} shows the results, and as the number of latents chosen increases, the average total change increases. We use $k = 50$ in our study.

\begin{table}[h]
\caption{\texttt{ValSel} average total change with varying top-k}
\centering
\small
\begin{tabular}{lccccccccc}
\textbf{Top-k} & 10 & 20 & 30 & 40 & 50 & 75 & 100 & 150 & 200 \\
\hline
Latent count & 10 & 20 & 30 & 40 & 50 & 75 & 100 & 150 & 200 \\
Avg total change & 8.75 & 10.84 & 11.24 & 11.40 & 11.23 & 14.11 & 14.22 & 14.26 & \textbf{15.36} \\
\hline
\end{tabular}
\label{tab:valsel-top-k}
\end{table}

\subsection{\texttt{FreqSel} sensivitity analysis}

\texttt{FreqSel} has three hyperparameters: cross-lingual threshold $T$, active example percentage $N$, and token threshold $M$, where $N$ and $M$ are expressed as a percentage. The default values are $T = 0.8$, $N = 98$, and $M = 10$. \Cref{tab:cross-lingual-threshold,tab:active-example-percentage,tab:token-threshold} show the sensitivity analysis results. When changing one hyperparameter, we use the default values for other hyperparameters.

\begin{table}[h]
\caption{\texttt{FreqSel} average total change with varying cross lingual threshold (T)}
\centering
\small
\begin{tabular}{lcccccccc}
\textbf{Cross lingual threshold (T)} & 0.6 & 0.7 & 0.75 & 0.8 & 0.85 & 0.9 & 0.95 & 1.0 \\
\hline
Latent count & 1140 & 1359 & 1451 & 1586 & 1707 & 1825 & 2034 & 2269 \\
Avg total change & 14.25 & 14.56 & 14.47 & 14.82 & 15.16 & 15.14 & \textbf{15.38} & 15.32 \\
\hline
\end{tabular}
\label{tab:cross-lingual-threshold}
\end{table}

\begin{table}[h]
\caption{\texttt{FreqSel} average total change with varying active example percentage threshold (N)}
\centering
\small
\begin{tabular}{lccccc}
\textbf{Active example percentage threshold (N)} & 80 & 85 & 90 & 95 & 98 \\
\hline
Latent count & 2928 & 2861 & 2721 & 2303 & 1586 \\
Avg total change & 14.22 & 14.22 & 14.16 & 14.40 & \textbf{14.82} \\
\hline
\end{tabular}
\label{tab:active-example-percentage}
\end{table}

\begin{table}[h]
\caption{\texttt{FreqSel} average total change with varying active token percentage threshold (M)}
\centering
\small
\begin{tabular}{lccccccc}
\textbf{Active token percentage threshold (M)} & 5 & 7.5 & 10 & 12.5 & 15 & 20 & 25 \\
\hline
Latent count & 1679 & 1645 & 1586 & 1465 & 1297 & 1022 & 818 \\
Avg total change & \textbf{15.74} & 15.64 & 14.83 & 14.44 & 14.07 & 12.75 & 12.32 \\
\hline
\end{tabular}
\label{tab:token-threshold}
\end{table}

\subsection{\texttt{AnnSel} sensivity analysis}

\texttt{AnnSel} has three phases: attribution graph construction, path pruning, and annotation-based latent selection as explained in \Cref{app:annsel_implementation}. Attribution graph construction phase has two hyperparameters: node threshold and edge threshold. The default values are $\texttt{node\_threshold} = 0.8$ and $\texttt{edge\_threshold} = 0.98$, and the results are shown in \Cref{tab:node-threshold,tab:edge-threshold}. We report the hyperparameter sweep on three hyperparameters in path pruning phase: first edge weight threshold, last edge weight threshold, and edge throughput threshold. The default values are $\texttt{first\_edge\_weight\_threshold} = 0.5$, $\texttt{last\_edge\_weight\_threshold} = 0.25$, and $\texttt{edge\_throughput\_threshold} = 0.1$. \Cref{tab:first-edge-weight-threshold,tab:last-edge-weight-threshold,tab:edge-throughput-threshold} show the results. The selection phase has one hyperparameter: frequency separation threshold, which defaults to 0.1. \Cref{tab:frequency-separation-threshold} shows the result. When doing a sweep on one hyperparameter, we use default values for all the other hyperparameters.

\begin{table}[h]
\caption{\texttt{AnnSel} average total change with varying node threshold}
\centering
\small
\begin{tabular}{lccccccc}
\textbf{Node threshold} & 0.6 & 0.7 & 0.75 & 0.8 & 0.85 & 0.9 & 0.95 \\
\hline
Latent count & 221 & 227 & 227 & 228 & 228 & 229 & 231 \\
Avg total change & 13.79 & 13.78 & 13.76 & 13.68 & 13.70 & \textbf{13.84} & 13.81 \\
\hline
\end{tabular}
\label{tab:node-threshold}
\end{table}

\begin{table}[h]
\caption{\texttt{AnnSel} average total change with varying edge threshold}
\centering
\small
\begin{tabular}{lccccc}
\textbf{Edge threshold} & 0.9 & 0.92 & 0.95 & 0.98 & 0.99 \\
\hline
Latent count & 228 & 229 & 228 & 228 & 227 \\
Avg total change & 13.68 & 13.67 & 13.68 & 13.68 & \textbf{13.71} \\
\hline
\end{tabular}
\label{tab:edge-threshold}
\end{table}

\begin{table}[h]
\caption{\texttt{AnnSel} average total change with varying first edge weight threshold}
\centering
\small
\begin{tabular}{lccccc}
\textbf{First edge weight threshold} & 0.3 & 0.4 & 0.5 & 0.6 & 0.7 \\
\hline
Latent count & 230 & 229 & 228 & 226 & 217 \\
Avg total change & 13.68 & 13.67 & 13.68 & \textbf{13.69} & 13.59 \\
\hline
\end{tabular}
\label{tab:first-edge-weight-threshold}
\end{table}

\begin{table}[h]
\caption{\texttt{AnnSel} average total change with varying last edge weight threshold}
\centering
\small
\begin{tabular}{lcccccc}
\textbf{Last edge weight threshold} & 0.1 & 0.15 & 0.2 & 0.25 & 0.3 & 0.35 \\
\hline
Latent count & 347 & 270 & 243 & 228 & 188 & 161 \\
Avg total change & \textbf{14.05} & 13.69 & 13.86 & 13.68 & 13.71 & 13.30 \\
\hline
\end{tabular}
\label{tab:last-edge-weight-threshold}
\end{table}

\begin{table}[h]
\caption{\texttt{AnnSel} average total change with varying edge throughput threshold}
\centering
\small
\begin{tabular}{lcccccc}
\textbf{Edge throughput threshold} & 0.05 & 0.075 & 0.1 & 0.15 & 0.2 & 0.25 \\
\hline
Latent count & 228 & 228 & 228 & 228 & 228 & 228 \\
Avg total change & 13.68 & 13.68 & 13.68 & 13.68 & 13.68 & 13.68 \\
\hline
\end{tabular}
\label{tab:edge-throughput-threshold}
\end{table}

\begin{table}[h]
\caption{\texttt{AnnSel} average total change with varying frequency separation threshold}
\centering
\small
\begin{tabular}{lcccccccccc}
\textbf{Frequency separation threshold} & 0.05 & 0.075 & 0.1 & 0.125 & 0.15 & 0.2 & 0.25 & 0.3 & 0.4 & 0.5 \\
\hline
Latent count & 564 & 324 & 229 & 178 & 139 & 92 & 71 & 59 & 42 & 35 \\
Avg total change & \textbf{15.73} & 14.85 & 13.74 & 13.42 & 12.91 & 11.77 & 11.01 & 9.94 & 9.40 & 9.24 \\
\hline
\end{tabular}
\label{tab:frequency-separation-threshold}
\end{table}

\section{Detailed Intervention Strategy Comparison}
Next, we provide detailed results for the different interventions under \texttt{Gemma-2-2B} and the \textbf{Antonyms} task in~\Cref{tab:int_V,tab:int_F,tab:int_A}. Similarly, the detailed results for different interventions under \texttt{Qwen3-4B} and the \textbf{Antonyms} task are shown in~\Cref{tab:intQ_V,tab:intQ_F,tab:intQ_A}.

\begin{table*}[!htp]\centering\small
\caption{Intervention effectiveness per target language for \texttt{ValSel} latents  of \texttt{Gemma-2-2B}}\label{tab:int_V}
% \resizebox{\textwidth}{!}{ % use this if the table is too large
\begin{tabular}{r|ccccccc}\toprule
 &Distractor ($l_0$)  &$L / \{l\}$ &Distractor ($l_0$)&Distractor ($l_0$)&Target  ($l$) &Zero+Amp &1L+Amp \\  
 
 & zero  & zero  &one-layer& multi-layer  & amplification &&\\  
 
 & ablation &ablation &direction ablation &direction ablation & && \\ \midrule
de &-3.18 &3.12 &-0.46 &-3.15 &14.52 &13.6 &13.92 \\
en &7.49 &5.06 &2.77 &52.07 &5.88 &10.16 &4.24 \\
es &0.25 &3.06 &-0.06 &-1.57 &13.35 &14.19 &15.08 \\
fr &-0.77 &1.15 &-0.28 &-9.65 &8.74 &9.01 &8.71 \\
ja &-4.61 &-1.57 &-0.41 &-11.59 &3.11 &4.62 &5.09 \\
ko &-0.08 &12.13 &-1.61 &-9.4 &10.38 &12.19 &7.58 \\
zh &-7.12 &-0.56 &2.04 &-2.73 &13.59 &16.06 &15.11 \\ \midrule
Total &-8.02 &22.39 &1.99 &13.98 &69.57 &\textbf{79.83} &\underline{69.73} \\
\bottomrule
\end{tabular}
% }
\end{table*}

\begin{table*}[!htp]\centering\small
\caption{Intervention effectiveness per target language for \texttt{FreqSel} latents  of \texttt{Gemma-2-2B}}\label{tab:int_F}
%\resizebox{\textwidth}{!}{ % use this if the table is too large
\begin{tabular}{r|ccccccc}\toprule
 &Distractor ($l_0$)  &$L / \{l\}$ &Distractor ($l_0$)&Distractor ($l_0$)&Target  ($l$) &Zero+Amp &1L+Amp \\  
 
 & zero  & zero  &one-layer& multi-layer  & amplification &&\\  
 
 & ablation &ablation &direction ablation &direction ablation & && \\ \midrule
de &-3.45 &0.8 &-1.67 &0.58 &5.08 &7.86 &6.29 \\
en &5.86 &5.43 &3.6 &-0.21 &-0.55 &4.36 &1.94 \\
es &-0.7 &3.39 &-0.6 &-5.12 &6.68 &8.95 &7.09 \\
fr &-1.51 &0.65 &-0.78 &-5.33 &4.45 &6.45 &5.09 \\
ja &-0.97 &1.38 &-1.45 &-27.91 &8.65 &8.65 &9.49 \\
ko &4.82 &6.62 &-3.3 &0.06 &29.54 &28.31 &27.6 \\
zh &-2.03 &3.12 &1.25 &-13.69 &36.98 &34.49 &38.66 \\ \midrule
Total &2.02 &21.39 &-2.95 &-51.62 &90.83 &\textbf{99.07} &\underline{96.16} \\
\bottomrule
\end{tabular}
\end{table*}

\begin{table*}[!htp]\centering\small
\caption{Intervention effectiveness per target language for \texttt{AnnSel} latents  of \texttt{Gemma-2-2B}}\label{tab:int_A}
%\resizebox{\textwidth}{!}{ % use this if the table is too large
\begin{tabular}{r|ccccccc}\toprule
 &Distractor ($l_0$)  &$L / \{l\}$ &Distractor ($l_0$)&Distractor ($l_0$)&Target  ($l$) &Zero+Amp &1L+Amp \\  
 
 & zero  & zero  &one-layer& multi-layer  & amplification &&\\  
 
 & ablation &ablation &direction ablation &direction ablation & && \\ \midrule
de &-3.87 &0.68 &-0.04 &-3.62 &20.15 &19.68 &20.6 \\
en &4.35 &4.83 &0.82 &43.07 &1.12 &4.88 &1.74 \\
es &-0.04 &1.98 &-0.15 &6.6 &13.91 &13.86 &14.05 \\
fr &-1.01 &0.62 &0.02 &-3.71 &13.69 &13.54 &13.95 \\
ja &-2.21 &1.5 &-1.01 &-4.97 &11.17 &11.12 &10.53 \\
ko &6.76 &5.97 &0.44 &12.19 &4.27 &9.13 &4.81 \\
zh &-4.18 &-3.68 &0.22 &14.49 &22.69 &24.51 &23.65 \\ \midrule
Total &-0.20 &11.90 &0.30 &64.05 &87.00 &\textbf{96.72 }&\underline{89.33} \\
\bottomrule
\end{tabular}
\end{table*}

\begin{table*}[!htp]\centering\small
\caption{Intervention effectiveness per target language for \texttt{ValSel} latents  of \texttt{Qwen3-4B}}\label{tab:intQ_V}
% \resizebox{\textwidth}{!}{ % use this if the table is too large
\begin{tabular}{r|ccccccc}\toprule
 &Distractor ($l_0$)  &$L / \{l\}$ &Distractor ($l_0$)&Distractor ($l_0$)&Target  ($l$) &Zero+Amp &1L+Amp \\  
 
 & zero  & zero  &one-layer& multi-layer  & amplification &&\\  
 
 & ablation &ablation &direction ablation &direction ablation & && \\ \midrule
de &5.24 &0.83 &-4.32 &-21.71 &4.58 &8.47 &4.99 \\
en &1.05 &6.6 &-2.81 &2.16 &-14.3 &-6.17 &-14.71 \\
es &3.35 &1.7 &-1.4 &-20.7 &2.79 &7.4 &4.44 \\
fr &3.51 &-0.23 &-2.79 &-16.11 &2.85 &7.51 &4.05 \\
ja &3.33 &-4.71 &-6.52 &-19.02 &-6.02 &-1.2 &-4.12 \\
ko &18.59 &2.5 &-2.11 &-5.64 &3.44 &11.81 &7.29 \\
zh &5.47 &8.84 &2.67 &17.1 &7.51 &17.74 &13.9 \\ \midrule
Total &\underline{40.54} &15.53 &-17.28 &-63.92 &0.85 &\textbf{45.56} &15.84 \\
\bottomrule
\end{tabular}
% }
\end{table*}

\begin{table*}[!htp]\centering\small
\caption{Intervention effectiveness per target language for \texttt{FreqSel} latents  of \texttt{Qwen3-4B}}\label{tab:intQ_F}
%\resizebox{\textwidth}{!}{ % use this if the table is too large
\begin{tabular}{r|ccccccc}\toprule
 &Distractor ($l_0$)  &$L / \{l\}$ &Distractor ($l_0$)&Distractor ($l_0$)&Target  ($l$) &Zero+Amp &1L+Amp \\  
 
 & zero  & zero  &one-layer& multi-layer  & amplification &&\\  
 
 & ablation &ablation &direction ablation &direction ablation & && \\ \midrule
de &-2.92 &0.15 &-0.28 &-24.44 &4.69 &6.35 &5.08 \\
en &6.67 &8.39 &1.19 &-9.42 &-2.01 &6.11 &-1.55 \\
es &1.41 &3.67 &0.35 &-21.37 &2.85 &4.92 &2.93 \\
fr &-0.96 &2.51 &-0.3 &-16.97 &5.05 &6.88 &5.18 \\
ja &-3.3 &0.18 &-3.63 &-37.57 &-0.27 &0.37 &-1.25 \\
ko &3.99 &7.29 &-0.25 &-30.15 &19.97 &21.29 &20.5 \\
zh &11.66 &11.86 &7.94 &-25.09 &26.47 &29.76 &27.71 \\ \midrule
Total &16.55 &34.05 &5.02 &-165.01 &56.75 &\textbf{75.68} &\underline{58.60} \\
\bottomrule
\end{tabular}
\end{table*}

\begin{table*}[!htp]\centering\small
\caption{Intervention effectiveness per target language for \texttt{AnnSel} latents  of \texttt{Qwen3-4B}}\label{tab:intQ_A}
%\resizebox{\textwidth}{!}{ % use this if the table is too large
\begin{tabular}{r|ccccccc}\toprule
 &Distractor ($l_0$)  &$L / \{l\}$ &Distractor ($l_0$)&Distractor ($l_0$)&Target  ($l$) &Zero+Amp &1L+Amp \\  
 
 & zero  & zero  &one-layer& multi-layer  & amplification &&\\  
 
 & ablation &ablation &direction ablation &direction ablation & && \\ \midrule
de &-0.67 &0 &0.02 &-2.51 &6.73 &6.49 &6.89 \\
en &4.62 &4.9 &0.6 &8.84 &0 &0 &0 \\
es &1.2 &0 &0.04 &-2.58 &6.31 &7.28 &6.41 \\
fr &-0.71 &0 &-0.02 &-6.21 &6.5 &6.35 &6.56 \\
ja &0.4 &0 &-0.18 &-13.1 &1.76 &1.83 &1.72 \\
ko &0.12 &0 &-2.42 &-5 &5.6 &7.7 &6.08 \\
zh &3.46 &0 &3.08 &-0.4 &6.39 &9.78 &9.73 \\ \midrule
Total &8.42 &4.9 &1.12 &-20.96 &33.29 &\textbf{39.43} & \underline{37.39} \\
\bottomrule
\end{tabular}
\end{table*}

\section{\textbf{Zero+Amp} Detailed Results}
\Cref{tab:valsel_logits,tab:freqsel_logits,tab:annsel_logits} show the logit difference results of \texttt{ValSel}, \texttt{FreqSel} and \texttt{AnnSel}, respectively, before and after the \textbf{Zero+Amp} intervention on \textbf{Antonyms} in \texttt{Gemma-2-2B}. The numbers in the table are for \textit{top-1 logit margin} to the target adjective, which is the separation in the logit space of the target token to its best competitor. A value $\le 0$ (green) indicates that the target adjective is the most likely token, while a positive value (red) means it lags behind other undesired tokens. 

Specifically, rows show the context language and columns show the adjective language. For each language pair, the `before' value shows the logit margin before any intervention, while `after' is its value after the latent amplification is applied. The desired outcome is that the value becomes smaller (greener) after intervention. Red cells indicate failures.

While the three methods show reasonable efficacy in enforcing Spanish and Chinese, \texttt{FreqSel} is weaker in enforcing German and English, and \texttt{ValSel} shows fragility when the context language is Chinese (Table~\ref{tab:valsel_logits}). The situation of French adjectives is unique, where \texttt{ValSel} and \texttt{FreqSel} fail for some context languages while \texttt{AnnSel} shows better robustness. On the other hand, \texttt{AnnSel} shows fragility in Korean, where only 7 latents were identified (Table~\ref{tab:latent_counts}) --- far fewer than for any other language, while \texttt{FreqSel} shows the most robust performance.

\begin{table*}[!htp]\centering
\caption{Logit difference for \texttt{ValSel} and  \texttt{Gemma-2-2B} before and after the  intervention  using \textbf{Antonyms}}\label{tab:valsel_logits}
\resizebox{\textwidth}{!}{ 
\begin{tabular}{lrrrrrrrrrrrrrrr}\toprule
&\multicolumn{2}{c}{es} &\multicolumn{2}{c}{en} &\multicolumn{2}{c}{zh} &\multicolumn{2}{c}{de} &\multicolumn{2}{c}{ja} &\multicolumn{2}{c}{fr} &\multicolumn{2}{c}{ko} \\\cmidrule{2-15}
&before &after &before &after &before &after &before &after &before &after &before &after &before &after \\\midrule

de &\cellcolor[HTML]{daf0e5}-1.79 &\cellcolor[HTML]{c4e7d5}-2.93 &\cellcolor[HTML]{a3d9be}-4.57 &\cellcolor[HTML]{96d4b6}-5.19 &\cellcolor[HTML]{c3e6d5}-2.97 &\cellcolor[HTML]{9cd7ba}-4.88 &- & - &\cellcolor[HTML]{cae9da}-2.62 &\cellcolor[HTML]{c8e9d9}-2.69 &\cellcolor[HTML]{fafdfb}-0.24 &\cellcolor[HTML]{ebf7f1}-0.98 &\cellcolor[HTML]{cfebdd}-2.38 &\cellcolor[HTML]{d2ede0}-2.21 \\
en &\cellcolor[HTML]{d0ecde}-2.29 &\cellcolor[HTML]{bce4d0}-3.29 &- & - &\cellcolor[HTML]{b3e0ca}-3.77 &\cellcolor[HTML]{a6dbc1}-4.41 &\cellcolor[HTML]{e1f3ea}-1.45 &\cellcolor[HTML]{cfebde}-2.35 &\cellcolor[HTML]{c3e6d5}-2.96 &\cellcolor[HTML]{c8e8d9}-2.71 &\cellcolor[HTML]{fcfefd}-0.10 &\cellcolor[HTML]{f1f9f5}-0.66 &\cellcolor[HTML]{c5e7d6}-2.87 &\cellcolor[HTML]{bde4d1}-3.27 \\
es &- & - &\cellcolor[HTML]{abddc4}-4.15 &\cellcolor[HTML]{89cfad}-5.83 &\cellcolor[HTML]{b3e0ca}-3.77 &\cellcolor[HTML]{a3d9bf}-4.56 &\cellcolor[HTML]{e4f4ec}-1.30 &\cellcolor[HTML]{c9e9d9}-2.68 &\cellcolor[HTML]{d1ecdf}-2.27 &\cellcolor[HTML]{cbe9da}-2.58 &\cellcolor[HTML]{fdfefd}-0.08 &\cellcolor[HTML]{d6eee2}-2.03 &\cellcolor[HTML]{d0ecde}-2.30 &\cellcolor[HTML]{d1ecdf}-2.25 \\
fr &\cellcolor[HTML]{e6f5ed}-1.22 &\cellcolor[HTML]{bee5d2}-3.19 &\cellcolor[HTML]{abddc5}-4.13 &\cellcolor[HTML]{98d5b7}-5.08 &\cellcolor[HTML]{d8efe4}-1.89 &\cellcolor[HTML]{aadcc3}-4.22 &\cellcolor[HTML]{eaf6f0}-1.02 &\cellcolor[HTML]{d1ecdf}-2.26 &\cellcolor[HTML]{cdeadc}-2.48 &\cellcolor[HTML]{c9e9d9}-2.65 &- & - &\cellcolor[HTML]{d2ecdf}-2.22 &\cellcolor[HTML]{d3ede0}-2.18 \\
ja &\cellcolor[HTML]{fbeae9}0.95 &\cellcolor[HTML]{d9efe4}-1.87 &\cellcolor[HTML]{b7e2cd}-3.54 &\cellcolor[HTML]{88ceac}-5.90 &\cellcolor[HTML]{fef8f7}0.34 &\cellcolor[HTML]{a5dac0}-4.45 &\cellcolor[HTML]{fcecea}0.87 &\cellcolor[HTML]{dcf0e6}-1.73 &- & - &\cellcolor[HTML]{f8dad8}1.65 &\cellcolor[HTML]{fefaf9}0.25 &\cellcolor[HTML]{f8d9d7}1.69 &\cellcolor[HTML]{bce4d0}-3.29 \\
ko &\cellcolor[HTML]{f9ddda}1.54 &\cellcolor[HTML]{cfebdd}-2.37 &\cellcolor[HTML]{bde4d1}-3.28 &\cellcolor[HTML]{a6dbc1}-4.40 &\cellcolor[HTML]{f6fbf8}-0.44 &\cellcolor[HTML]{85cdaa}-6.04 &\cellcolor[HTML]{fcedec}0.80 &\cellcolor[HTML]{cceadb}-2.53 &\cellcolor[HTML]{d7eee3}-1.98 &\cellcolor[HTML]{cbeadb}-2.54 &\cellcolor[HTML]{f8d8d5}1.75 &\cellcolor[HTML]{f2faf6}-0.61 &- & - \\
zh &\cellcolor[HTML]{f4c4c0}2.62 &\cellcolor[HTML]{f0f9f4}-0.73 &\cellcolor[HTML]{d0ecde}-2.29 &\cellcolor[HTML]{8bd0ae}-5.72 &- & - &\cellcolor[HTML]{f1b3ae}3.38 &\cellcolor[HTML]{eff8f4}-0.77 &\cellcolor[HTML]{f6cdc9}2.25 &\cellcolor[HTML]{e0f2e9}-1.52 &\cellcolor[HTML]{f3c0bc}2.80 &\cellcolor[HTML]{fcedec}0.80 &\cellcolor[HTML]{e67c73}5.79 &\cellcolor[HTML]{e5f4ed}-1.27 \\
\midrule
Total &-0.19 &-14.38 &-21.96 &-32.12 &-12.50 &-28.56 &1.28 &-12.32 &-10.06 &-14.69 &5.78 &-3.23 &-2.29 &-14.47 \\

Change &&+14.19&& +10.16&& +16.06&& +13.6&& +4.63&& +9.01&& +12.18\\

\bottomrule
\end{tabular}
}
\end{table*}

\begin{table*}[!htp]\centering
\caption{Logit difference for \texttt{FreqSel}  and \texttt{Gemma-2-2B} before and after the  intervention using \textbf{Antonyms}}\label{tab:freqsel_logits}
\resizebox{\textwidth}{!}{ 
\begin{tabular}{lrrrrrrrrrrrrrrr}\toprule
&\multicolumn{2}{c}{es} &\multicolumn{2}{c}{en} &\multicolumn{2}{c}{zh} &\multicolumn{2}{c}{de} &\multicolumn{2}{c}{ja} &\multicolumn{2}{c}{fr} &\multicolumn{2}{c}{ko} \\\cmidrule{2-15}
&before &after &before &after &before &after &before &after &before &after &before &after &before &after \\\midrule

de &\cellcolor[HTML]{daf0e5}-1.79 &\cellcolor[HTML]{dff2e8}-1.59 &\cellcolor[HTML]{a3d9be}-4.57 &\cellcolor[HTML]{9bd6b9}-4.97 &\cellcolor[HTML]{c3e6d5}-2.97 &\cellcolor[HTML]{66c194}-7.58 &- & - &\cellcolor[HTML]{cae9da}-2.62 &\cellcolor[HTML]{c0e5d3}-3.13 &\cellcolor[HTML]{fafdfb}-0.24 &\cellcolor[HTML]{fbfdfc}-0.18 &\cellcolor[HTML]{cfebdd}-2.38 &\cellcolor[HTML]{89cfad}-5.82 \\
en &\cellcolor[HTML]{d0ecde}-2.29 &\cellcolor[HTML]{dff2e9}-1.55 &- & - &\cellcolor[HTML]{b3e0ca}-3.77 &\cellcolor[HTML]{5ebd8f}-7.99 &\cellcolor[HTML]{e1f3ea}-1.45 &\cellcolor[HTML]{e7f5ee}-1.17 &\cellcolor[HTML]{c3e6d5}-2.96 &\cellcolor[HTML]{bbe3d0}-3.34 &\cellcolor[HTML]{fcfefd}-0.10 &\cellcolor[HTML]{fffcfc}0.16 &\cellcolor[HTML]{c5e7d6}-2.87 &\cellcolor[HTML]{8bd0ae}-5.73 \\
es &- & - &\cellcolor[HTML]{abddc4}-4.15 &\cellcolor[HTML]{a2d9be}-4.62 &\cellcolor[HTML]{b3e0ca}-3.77 &\cellcolor[HTML]{59bb8b}-8.23 &\cellcolor[HTML]{e4f4ec}-1.30 &\cellcolor[HTML]{eaf6f0}-1.03 &\cellcolor[HTML]{d1ecdf}-2.27 &\cellcolor[HTML]{bee4d2}-3.21 &\cellcolor[HTML]{fdfefd}-0.08 &\cellcolor[HTML]{f9fcfb}-0.26 &\cellcolor[HTML]{d0ecde}-2.30 &\cellcolor[HTML]{8ed1b0}-5.61 \\
fr &\cellcolor[HTML]{e6f5ed}-1.22 &\cellcolor[HTML]{dbf0e6}-1.75 &\cellcolor[HTML]{abddc5}-4.13 &\cellcolor[HTML]{a0d8bd}-4.70 &\cellcolor[HTML]{d8efe4}-1.89 &\cellcolor[HTML]{62bf91}-7.79 &\cellcolor[HTML]{eaf6f0}-1.02 &\cellcolor[HTML]{eaf6f0}-1.04 &\cellcolor[HTML]{cdeadc}-2.48 &\cellcolor[HTML]{bce4d0}-3.30 &- & - &\cellcolor[HTML]{d2ecdf}-2.22 &\cellcolor[HTML]{88ceac}-5.90 \\
ja &\cellcolor[HTML]{fbeae9}0.95 &\cellcolor[HTML]{e1f3ea}-1.45 &\cellcolor[HTML]{b7e2cd}-3.54 &\cellcolor[HTML]{adddc6}-4.06 &\cellcolor[HTML]{fef8f7}0.34 &\cellcolor[HTML]{70c59c}-7.06 &\cellcolor[HTML]{fcecea}0.87 &\cellcolor[HTML]{e9f6f0}-1.05 &- & - &\cellcolor[HTML]{f8dad8}1.65 &\cellcolor[HTML]{fffbfb}0.19 &\cellcolor[HTML]{f8d9d7}1.69 &\cellcolor[HTML]{aedec7}-3.99 \\
ko &\cellcolor[HTML]{f9ddda}1.54 &\cellcolor[HTML]{d9efe5}-1.85 &\cellcolor[HTML]{bde4d1}-3.28 &\cellcolor[HTML]{a9dcc3}-4.26 &\cellcolor[HTML]{f6fbf8}-0.44 &\cellcolor[HTML]{57bb8a}-8.35 &\cellcolor[HTML]{fcedec}0.80 &\cellcolor[HTML]{e4f4ec}-1.30 &\cellcolor[HTML]{d7eee3}-1.98 &\cellcolor[HTML]{c0e5d3}-3.13 &\cellcolor[HTML]{f8d8d5}1.75 &\cellcolor[HTML]{f3faf6}-0.59 &- & - \\
zh &\cellcolor[HTML]{f4c4c0}2.62 &\cellcolor[HTML]{ebf7f1}-0.95 &\cellcolor[HTML]{d0ecde}-2.29 &\cellcolor[HTML]{b4e0cb}-3.69 &- & - &\cellcolor[HTML]{f1b3ae}3.38 &\cellcolor[HTML]{ebf7f1}-0.98 &\cellcolor[HTML]{f6cdc9}2.25 &\cellcolor[HTML]{cae9da}-2.60 &\cellcolor[HTML]{f3c0bc}2.80 &0.01 &\cellcolor[HTML]{e67c73}5.79 &\cellcolor[HTML]{b7e2cd}-3.54 \\
\midrule
Total &-0.19 &-9.14 &-21.96 &-26.30 &-12.50 &-47.00 &1.28 &-6.57 &-10.06 &-18.71 &5.78 &-0.67 &-2.29 &-30.59 \\
Change &&+8.95&& +4.34&& +34.5&& +7.85&& +8.65&& +6.45&& +28.3 \\
\bottomrule
\end{tabular}
}
\end{table*}

\begin{table*}[!htp]\centering
\caption{Logit difference for \texttt{AnnSel}  and \texttt{Gemma-2-2B} before and after the  intervention  using \textbf{Antonyms}}\label{tab:annsel_logits}
\resizebox{\textwidth}{!}{ 
\begin{tabular}{lrrrrrrrrrrrrrrr}\toprule
&\multicolumn{2}{c}{es} &\multicolumn{2}{c}{en} &\multicolumn{2}{c}{zh} &\multicolumn{2}{c}{de} &\multicolumn{2}{c}{ja} &\multicolumn{2}{c}{fr} &\multicolumn{2}{c}{ko} \\\cmidrule{2-15}
&before &after &before &after &before &after &before &after &before &after &before &after &before &after \\\midrule

de &\cellcolor[HTML]{daf0e5}-1.79 &\cellcolor[HTML]{c1e6d4}-3.04 &\cellcolor[HTML]{a3d9be}-4.57 &\cellcolor[HTML]{93d3b4}-5.32 &\cellcolor[HTML]{c3e6d5}-2.97 &\cellcolor[HTML]{8cd0af}-5.67 &- & - &\cellcolor[HTML]{cae9da}-2.62 &\cellcolor[HTML]{b5e1cc}-3.63 &\cellcolor[HTML]{fafdfb}-0.24 &\cellcolor[HTML]{ddf1e7}-1.67 &\cellcolor[HTML]{cfebdd}-2.38 &\cellcolor[HTML]{cbeadb}-2.54 \\
en &\cellcolor[HTML]{d0ecde}-2.29 &\cellcolor[HTML]{c0e5d3}-3.11 &- & - &\cellcolor[HTML]{b3e0ca}-3.77 &\cellcolor[HTML]{7dcaa4}-6.46 &\cellcolor[HTML]{e1f3ea}-1.45 &\cellcolor[HTML]{c2e6d4}-3.01 &\cellcolor[HTML]{c3e6d5}-2.96 &\cellcolor[HTML]{b4e0cb}-3.69 &\cellcolor[HTML]{fcfefd}-0.10 &\cellcolor[HTML]{eaf6f0}-1.04 &\cellcolor[HTML]{c5e7d6}-2.87 &\cellcolor[HTML]{cbeada}-2.57 \\
es &- & - &\cellcolor[HTML]{abddc4}-4.15 &\cellcolor[HTML]{9dd7bb}-4.84 &\cellcolor[HTML]{b3e0ca}-3.77 &\cellcolor[HTML]{7bc9a3}-6.52 &\cellcolor[HTML]{e4f4ec}-1.30 &\cellcolor[HTML]{bbe3cf}-3.37 &\cellcolor[HTML]{d1ecdf}-2.27 &\cellcolor[HTML]{b6e1cc}-3.62 &\cellcolor[HTML]{fdfefd}-0.08 &\cellcolor[HTML]{d1ecdf}-2.26 &\cellcolor[HTML]{d0ecde}-2.30 &\cellcolor[HTML]{cceadb}-2.50 \\
fr &\cellcolor[HTML]{e6f5ed}-1.22 &\cellcolor[HTML]{c1e6d4}-3.06 &\cellcolor[HTML]{abddc5}-4.13 &\cellcolor[HTML]{9cd7ba}-4.88 &\cellcolor[HTML]{d8efe4}-1.89 &\cellcolor[HTML]{88ceac}-5.91 &\cellcolor[HTML]{eaf6f0}-1.02 &\cellcolor[HTML]{bee4d1}-3.23 &\cellcolor[HTML]{cdeadc}-2.48 &\cellcolor[HTML]{b3e0ca}-3.74 &- & - &\cellcolor[HTML]{d2ecdf}-2.22 &\cellcolor[HTML]{cceadc}-2.49 \\
ja &\cellcolor[HTML]{fbeae9}0.95 &\cellcolor[HTML]{def1e8}-1.62 &\cellcolor[HTML]{b7e2cd}-3.54 &\cellcolor[HTML]{a0d8bd}-4.69 &\cellcolor[HTML]{fef8f7}0.34 &\cellcolor[HTML]{86ceaa}-6.00 &\cellcolor[HTML]{fcecea}0.87 &\cellcolor[HTML]{c8e8d9}-2.71 &- & - &\cellcolor[HTML]{f8dad8}1.65 &\cellcolor[HTML]{f3faf6}-0.58 &\cellcolor[HTML]{f8d9d7}1.69 &\cellcolor[HTML]{d5eee1}-2.07 \\
ko &\cellcolor[HTML]{f9ddda}1.54 &\cellcolor[HTML]{cfebdd}-2.36 &\cellcolor[HTML]{bde4d1}-3.28 &\cellcolor[HTML]{bce3d0}-3.33 &\cellcolor[HTML]{f6fbf8}-0.44 &\cellcolor[HTML]{7dcaa4}-6.46 &\cellcolor[HTML]{fcedec}0.80 &\cellcolor[HTML]{b7e2cd}-3.55 &\cellcolor[HTML]{d7eee3}-1.98 &\cellcolor[HTML]{b7e2cd}-3.53 &\cellcolor[HTML]{f8d8d5}1.75 &\cellcolor[HTML]{def1e8}-1.61 &- & - \\
zh &\cellcolor[HTML]{f4c4c0}2.62 &\cellcolor[HTML]{edf7f2}-0.86 &\cellcolor[HTML]{d0ecde}-2.29 &\cellcolor[HTML]{b3e0ca}-3.77 &- & - &\cellcolor[HTML]{f1b3ae}3.38 &\cellcolor[HTML]{cceadb}-2.52 &\cellcolor[HTML]{f6cdc9}2.25 &\cellcolor[HTML]{c3e6d5}-2.97 &\cellcolor[HTML]{f3c0bc}2.80 &\cellcolor[HTML]{f2faf6}-0.60 &\cellcolor[HTML]{e67c73}5.79 &\cellcolor[HTML]{fceeed}0.77 \\
\midrule
Total &-0.19 &-14.05 &-21.96 &-26.83 &-12.50 &-37.02 &1.28 &-18.39 &-10.06 &-21.18 &5.78 &-7.76 &-2.29 &-11.40 \\

Change && +13.86&& +4.87&& +24.52&& +19.67&& +11.12&& +13.54&& +9.11 \\
\bottomrule
\end{tabular}
}
\end{table*}

\section{Detailed Selection Method Comparison}

\Cref{tab:total_logits_qwen} presents summary results for selection method efficacy for \texttt{Qwen3-4B} latents for \textbf{Antonyms} and \textbf{Enumerations}.

\begin{table}[ht!]
    \caption{Selection method efficacy per language under the \textbf{Zero+Amp} intervention for \texttt{Qwen3-4B}. Post-intervention total change in top-1 logit margin for
\textbf{Antonyms}; Bottom: average change in normalized logprob of the target sequence for \textbf{Enumerations}.
}
\begin{center}\resizebox{0.65\textwidth}{!}{\begin{tabular}{r|r|ccccccc|c}
    \toprule
        &&es&en&zh&de& ja&fr&ko & Avg\\
        \hline
\multirow{3}{.4cm}{\rotatebox[origin=c]{90}{\textbf{Antons.}}}

    % \centering\resizebox{0.5\textwidth}{!}{ 
    % \begin{tabular}{r|ccccccc|c}
    % \toprule
    %     &es&en&zh&de& ja&fr&ko & Avg\\
    %     \hline
&\texttt{ValSel} &\textbf{7.36} &-5.31 &\underline{17.56} &\textbf{6.38} &\underline{-1.17} &\textbf{7.49} &\underline{11.29} &\underline{6.23} \\
    &\texttt{FreqSel} &4.90 &\textbf{6.09} &\textbf{29.87} &4.13 &-7.34 &6.90 &\textbf{20.86} &\textbf{9.34} \\
&\texttt{AnnSel} &\underline{7.35} &\underline{5.12} &9.84 &\underline{4.95} &\textbf{2.23} &\underline{6.95} &6.79 &6.17 \\
 \bottomrule    
 % \toprule    
\multirow{3}{.4cm}{\rotatebox[origin=c]{90}{\textbf{Enums.}}}
&\texttt{ValSel} & \underline{0.72}&{0.00} &\underline{0.19} &\underline{0.78} &\underline{0.30} & \underline{0.79}& \underline{0.40}& \underline{0.45}\\
    &\texttt{FreqSel} & \textbf{1.20}& \textbf{0.27}& \textbf{1.09}& \textbf{1.36}& \textbf{0.90}& \textbf{1.23}& \textbf{1.27}& \textbf{1.05}\\
&\texttt{AnnSel} & 0.51& \underline{0.17} & 0.10& 0.61& 0.12& 0.64& 0.35& 0.36\\
\bottomrule        
    \end{tabular}
   }

    \label{tab:total_logits_qwen}
    \end{center}
\end{table}

\section{Continued Generation}
\label{app:cont_generation}

To verify whether the latents we found are effective outside our evaluation datasets and in real-life generation, we take sentences from FLORES+~\citep{nllb-24} that we do not use when finding the latents and do top-p decoding with $p=0.5$ while applying Zero+Amp intervention. We show some examples of the generated sentences in \Cref{tab:continuation}. For the full list of generated sentences, please refer to the GitHub repo.

\begin{table}[t]
\centering
\footnotesize
\caption{Selected held-out FLORES+ continuations under feature intervention (ablate $A$, amplify $B$, $A\neq B$). Gemma-2-2B, default latent sets, top-$p=0.5$. The generated column is the model suffix after the midpoint-truncated prompt.}
\label{tab:flores-good-cross-lang-p05}
\begin{tabularx}{\textwidth}{l c c X X}
\toprule
Method & $A$ & $B$ & Original sentence (language $A$) & Generated continuation (language $B$) \\
\midrule
AnnSel & es & fr
  & La cantidad de hablantes nativos ser\'a diferente seg\'un se computen o no dialectos que est\'en muy relacionados entre s\'i.
  & \ldots\ situe la langue dans l'\'echelle des langues de la plan\`ete. \\
\addlinespace
ValSel & fr & es
  & Le nombre de langues maternelles d\'epend de si oui ou non les dialectes tr\`es proches sont pris en compte.
  & \ldots, si la lengua materna es una lengua de la familia o si es una lengua no de la familia. \\
\addlinespace
FreqSel & fr & es
  & Le nombre de langues maternelles d\'epend de si oui ou non les dialectes tr\`es proches sont pris en compte.
  & \ldots\ es la \'unica lengua materna o si hay otras. \\
\addlinespace
AnnSel & de & es
  & Die Anzahl der Muttersprachler variiert abh\"angig davon, ob sehr eng verwandte Dialekte mitgez\"ahlt werden oder nicht.
  & \ldots\ en ingl\'es o un espa\~nol es la lengua materna de los estudiantes. \\
\addlinespace
AnnSel & zh & es
  & \begin{CJK}{UTF8}{gbsn}母语人数取决于紧密相关的方言是否计算在内。\end{CJK}
  & \ldots mente de los diferentes factores que afectan la cantidad de hablantes de un idioma. \\
\addlinespace
FreqSel & ja & fr
  & \begin{CJK}{UTF8}{min}母語話者の人数は、言語的に非常に近い方言を数に入れるかどうかによって変わってきます。\end{CJK}
  & \ldots\ diversifiez la culture de votre enfant. \\
\addlinespace
AnnSel & fr & es
  & Vous vous fatigueriez moins de cette fa\c{c}on. N'oubliez pas qu'il n'est pas n\'ecessaire de trop insister sur les touches pour qu'elles r\'esonnent plus fort. C'est le cas sur le piano.
  & \ldots\ trabajar a fondo en una jornada. \\
\addlinespace
AnnSel & ja & es
  & \begin{CJK}{UTF8}{min}そうすることで、疲れがたまりにくくなります。ピアノのように音量を出すために力を入れて鍵盤を叩く必要はないことを覚えておいてください。\end{CJK}
  & \ldots\ instrumentos musicales en casa que no requieren de mucha fuerza. \\
\addlinespace
AnnSel & ko & ja
  & \begin{CJK}{UTF8}{mj}이 방식을 통해 연주의 피로도를 최대한 낮출 수 있습니다. 피아노와 달리, 아코디언은 더 큰 소리를 내기 위해 건반을 세게 칠 필요가 없다는 것을 기억하세요.\end{CJK}
  & \ldots\ \begin{CJK}{UTF8}{min}3弦ヴァイオリンは2弦ヴァイオリンよりも弦の張力が強め、弦の弾き方にもよりますが、音響的にはより響きが良きです。\end{CJK} \\
\addlinespace
AnnSel & fr & es
  & Le rep\'erage et la collecte d'informations par David Headley ont facilit\'e l'op\'eration men\'ee par les dix hommes arm\'es du groupe militant pakistanais Laskhar-e-Taiba.
  & \ldots\'ee en 2008 en Pakistan, a relatado el ministro de Defensa de los Estados Unidos, Robert Gates, durante un discurso en el Colegio de la Guerra de Estados Unidos, citado por la agencia de noticias AFP. \\
\addlinespace
FreqSel & zh & en
  & \begin{CJK}{UTF8}{gbsn}他又称：“这个病例很严重。但我们的医疗系统运作良好，请大家放心。”\end{CJK}
  & \ldots\ fortunately, it's not as serious as we thought.'' \\
\addlinespace
AnnSel & ko & ja
  & ``\begin{CJK}{UTF8}{mj}그는 `이건 심각한 사건입니다. 다행히도 우리 시스템이 잘 작동하고 있으니 안심하셔도 됩니다.'라고 말을 이었다.\end{CJK}''
  & \ldots\ \begin{CJK}{UTF8}{min}誤作動しただけで、100\%確信し得る情報はありませんと語りました。サイバーセキュリティは、あらゆるものを守るための重要な手段です。\end{CJK} \\
\addlinespace
FreqSel & de & fr
  & Er fuhr fort: ,,Dieser Fall ist ernst. Seien Sie versichert, dass unser System so gut funktioniert, wie es sollte.``
  & \ldots\ tr\`es s\'erieux. \\
\bottomrule
\end{tabularx}
\label{tab:continuation}
\end{table}

\section{Adjectives Used in \textbf{Antonyms}}
\label{app:antonyms}
We prepare 100 pairs of antonyms to use for the antonym-structured sentences (\textit{The opposite of "{adj1}" is ", {adj2}}). Keep in mind that there could be multiple suitable words for \textit{adj2} (e.g., the antonym of small could be either big or large). A human annotator first compiled an initial set of 30 adjective antonym pairs in English. An LLM (Gemini 2.5 Flash~\citep{comanici2025gemini25pushingfrontier}) was then instructed to expand this list to 100 pairs, after which every pair was manually validated. The validated pairs were subsequently translated into the remaining six languages using an LLM, and native speakers verified the translations for their respective languages. All the antonyms are shown in \Cref{tab:antonyms}. It is worth mentioning that the words or tokens can be exactly the same in some cases across different languages, most commonly between French and Spanish and between Chinese and Japanese.

%For Antonyms, a human annotator first compiled an initial set of 30 adjective antonym pairs in English. An LLM was then instructed to expand this list to 100 pairs, after which every pair was manually validated. The validated pairs were subsequently translated into the remaining six languages using an LLM, and native speakers verified the translations for their respective languages. We use Gemini 2.5 Flash~\citep{comanici2025gemini25pushingfrontier} to generate them. 

\clearpage
\onecolumn
\footnotesize
\setlength{\tabcolsep}{3pt}
\begin{longtable}{p{2.1cm} p{2.2cm} p{2.5cm} p{2.5cm} p{2cm} p{3cm} p{2.3cm}}
\label{tab:antonyms} \\
\caption{Multilingual Antonym Pairs in the 7 languages} \\
\toprule
\textbf{English} & \textbf{French} & \textbf{German} & \textbf{Spanish} &  \textbf{Chinese} & 
\textbf{Japanese} &\textbf{Korean} \\
\midrule
\endfirsthead

\toprule
\textbf{English} & \textbf{French} & \textbf{German} & \textbf{Spanish} &  \textbf{Chinese} & 
\textbf{Japanese} &\textbf{Korean} \\ 
\midrule
\endhead

\bottomrule
\endlastfoot

\textbf{good} & \textbf{bon} & \textbf{gut} & \textbf{bueno} & \textbf{\begin{CJK}{UTF8}{gbsn}好\end{CJK}} & \textbf{\begin{CJK}{UTF8}{min}良い\end{CJK}} & \textbf{\begin{CJK}{UTF8}{mj}좋은\end{CJK}} \\*
\textit{bad} & \textit{mauvais} & \textit{schlecht} & \textit{malo} & \textit{\begin{CJK}{UTF8}{gbsn}坏\end{CJK}} & \textit{\begin{CJK}{UTF8}{min}悪い\end{CJK}} & \textit{\begin{CJK}{UTF8}{mj}나쁜\end{CJK}} \\ \midrule
\textbf{happy} & \textbf{heureux} & \textbf{glücklich} & \textbf{feliz} & \textbf{\begin{CJK}{UTF8}{gbsn}开心\end{CJK}} & \textbf{\begin{CJK}{UTF8}{min}嬉しい\end{CJK}} & \textbf{\begin{CJK}{UTF8}{mj}행복한\end{CJK}} \\*
\textit{sad, unhappy} & \textit{triste, malheureux} & \textit{traurig, unglücklich} & \textit{triste, infeliz} & \textit{\begin{CJK}{UTF8}{gbsn}难过, 不高兴\end{CJK}} & \textit{\begin{CJK}{UTF8}{min}悲しい\end{CJK}} & \textit{\begin{CJK}{UTF8}{mj}슬픈, 불행한\end{CJK}} \\ \midrule
\textbf{big} & \textbf{grand} & \textbf{groß} & \textbf{grande} & \textbf{\begin{CJK}{UTF8}{gbsn}大\end{CJK}} & \textbf{\begin{CJK}{UTF8}{min}大きい\end{CJK}} & \textbf{\begin{CJK}{UTF8}{mj}큰\end{CJK}} \\*
\textit{small} & \textit{petit} & \textit{klein} & \textit{pequeño} & \textit{\begin{CJK}{UTF8}{gbsn}小\end{CJK}} & \textit{\begin{CJK}{UTF8}{min}小さい\end{CJK}} & \textit{\begin{CJK}{UTF8}{mj}작은\end{CJK}} \\ \midrule
\textbf{hot} & \textbf{chaud} & \textbf{heiß} & \textbf{caliente} & \textbf{\begin{CJK}{UTF8}{gbsn}热\end{CJK}} & \textbf{\begin{CJK}{UTF8}{min}暑い\end{CJK}} & \textbf{\begin{CJK}{UTF8}{mj}더운\end{CJK}} \\*
\textit{cold} & \textit{froid} & \textit{kalt} & \textit{frío} & \textit{\begin{CJK}{UTF8}{gbsn}冷\end{CJK}} & \textit{\begin{CJK}{UTF8}{min}寒い, 冷たい\end{CJK}} & \textit{\begin{CJK}{UTF8}{mj}추운, 차가운\end{CJK}} \\ \midrule
\textbf{fast} & \textbf{rapide} & \textbf{schnell} & \textbf{rápido} & \textbf{\begin{CJK}{UTF8}{gbsn}快\end{CJK}} & \textbf{\begin{CJK}{UTF8}{min}速い\end{CJK}} & \textbf{\begin{CJK}{UTF8}{mj}빠른\end{CJK}} \\*
\textit{slow} & \textit{lent, lente} & \textit{langsam} & \textit{lento} & \textit{\begin{CJK}{UTF8}{gbsn}慢\end{CJK}} & \textit{\begin{CJK}{UTF8}{min}遅い\end{CJK}} & \textit{\begin{CJK}{UTF8}{mj}느린\end{CJK}} \\ \midrule
\textbf{light} & \textbf{léger} & \textbf{leicht} & \textbf{ligero} & \textbf{\begin{CJK}{UTF8}{gbsn}轻\end{CJK}} & \textbf{\begin{CJK}{UTF8}{min}軽い\end{CJK}} & \textbf{\begin{CJK}{UTF8}{mj}가벼운\end{CJK}} \\*
\textit{heavy, dark} & \textit{lourd} & \textit{schwer} & \textit{pesado} & \textit{\begin{CJK}{UTF8}{gbsn}重\end{CJK}} & \textit{\begin{CJK}{UTF8}{min}重い\end{CJK}} & \textit{\begin{CJK}{UTF8}{mj}무거운\end{CJK}} \\ \midrule
\textbf{easy} & \textbf{facile} & \textbf{einfach} & \textbf{fácil} & \textbf{\begin{CJK}{UTF8}{gbsn}容易\end{CJK}} & \textbf{\begin{CJK}{UTF8}{min}簡単な\end{CJK}} & \textbf{\begin{CJK}{UTF8}{mj}쉬운\end{CJK}} \\*
\textit{difficult, hard} & \textit{difficile} & \textit{schwierig, schwer} & \textit{difícil} & \textit{\begin{CJK}{UTF8}{gbsn}难\end{CJK}} & \textit{\begin{CJK}{UTF8}{min}難しい\end{CJK}} & \textit{\begin{CJK}{UTF8}{mj}어려운\end{CJK}} \\ \midrule
\textbf{new} & \textbf{nouveau} & \textbf{neu} & \textbf{nuevo} & \textbf{\begin{CJK}{UTF8}{gbsn}新\end{CJK}} & \textbf{\begin{CJK}{UTF8}{min}新しい\end{CJK}} & \textbf{\begin{CJK}{UTF8}{mj}새로운\end{CJK}} \\*
\textit{old} & \textit{vieux, ancien} & \textit{alt} & \textit{viejo} & \textit{\begin{CJK}{UTF8}{gbsn}旧\end{CJK}} & \textit{\begin{CJK}{UTF8}{min}古い\end{CJK}} & \textit{\begin{CJK}{UTF8}{mj}오래된\end{CJK}} \\ \midrule
\textbf{true} & \textbf{vrai} & \textbf{wahr} & \textbf{verdadero} & \textbf{\begin{CJK}{UTF8}{gbsn}真\end{CJK}} & \textbf{\begin{CJK}{UTF8}{min}正しい\end{CJK}} & \textbf{\begin{CJK}{UTF8}{mj}진실한\end{CJK}} \\*
\textit{false, untrue} & \textit{faux} & \textit{falsch, unwahr} & \textit{falso, incorrecto} & \textit{\begin{CJK}{UTF8}{gbsn}假\end{CJK}} & \textit{\begin{CJK}{UTF8}{min}間違った\end{CJK}} & \textit{\begin{CJK}{UTF8}{mj}거짓의, 틀린\end{CJK}} \\ \midrule
\textbf{alive} & \textbf{vivant} & \textbf{lebendig} & \textbf{vivo} & \textbf{\begin{CJK}{UTF8}{gbsn}活\end{CJK}} & \textbf{\begin{CJK}{UTF8}{min}生きている\end{CJK}} & \textbf{\begin{CJK}{UTF8}{mj}살아있는\end{CJK}} \\*
\textit{dead} & \textit{mort} & \textit{tot} & \textit{muerto} & \textit{\begin{CJK}{UTF8}{gbsn}死\end{CJK}} & \textit{\begin{CJK}{UTF8}{min}死んでいる\end{CJK}} & \textit{\begin{CJK}{UTF8}{mj}죽은\end{CJK}} \\ \midrule
\textbf{full} & \textbf{plein} & \textbf{voll} & \textbf{lleno} & \textbf{\begin{CJK}{UTF8}{gbsn}满\end{CJK}} & \textbf{\begin{CJK}{UTF8}{min}満杯\end{CJK}} & \textbf{\begin{CJK}{UTF8}{mj}가득 찬\end{CJK}} \\*
\textit{empty} & \textit{vide} & \textit{leer} & \textit{vacío} & \textit{\begin{CJK}{UTF8}{gbsn}空\end{CJK}} & \textit{\begin{CJK}{UTF8}{min}空っぽ\end{CJK}} & \textit{\begin{CJK}{UTF8}{mj}빈\end{CJK}} \\ \midrule
\textbf{bright} & \textbf{brillant} & \textbf{hell} & \textbf{brillante} & \textbf{\begin{CJK}{UTF8}{gbsn}亮\end{CJK}} & \textbf{\begin{CJK}{UTF8}{min}明るい\end{CJK}} & \textbf{\begin{CJK}{UTF8}{mj}밝은\end{CJK}} \\*
\textit{dark, dim} & \textit{sombre, obscur} & \textit{dunkel} & \textit{oscuro, opaco} & \textit{\begin{CJK}{UTF8}{gbsn}暗\end{CJK}} & \textit{\begin{CJK}{UTF8}{min}暗い\end{CJK}} & \textit{\begin{CJK}{UTF8}{mj}어두운, 흐릿한\end{CJK}} \\ \midrule
\textbf{strong} & \textbf{fort} & \textbf{stark} & \textbf{fuerte} & \textbf{\begin{CJK}{UTF8}{gbsn}强\end{CJK}} & \textbf{\begin{CJK}{UTF8}{min}強い\end{CJK}} & \textbf{\begin{CJK}{UTF8}{mj}강한\end{CJK}} \\*
\textit{weak} & \textit{faible} & \textit{schwach} & \textit{débil} & \textit{\begin{CJK}{UTF8}{gbsn}弱\end{CJK}} & \textit{\begin{CJK}{UTF8}{min}弱い\end{CJK}} & \textit{\begin{CJK}{UTF8}{mj}약한\end{CJK}} \\ \midrule
\textbf{clean} & \textbf{propre} & \textbf{sauber} & \textbf{limpio} & \textbf{\begin{CJK}{UTF8}{gbsn}干净\end{CJK}} & \textbf{\begin{CJK}{UTF8}{min}きれいな\end{CJK}} & \textbf{\begin{CJK}{UTF8}{mj}깨끗한\end{CJK}} \\*
\textit{dirty} & \textit{sale} & \textit{schmutzig} & \textit{sucio} & \textit{\begin{CJK}{UTF8}{gbsn}脏\end{CJK}} & \textit{\begin{CJK}{UTF8}{min}汚い\end{CJK}} & \textit{\begin{CJK}{UTF8}{mj}더러운\end{CJK}} \\ \midrule
\textbf{open} & \textbf{ouvert} & \textbf{offen} & \textbf{abierto} & \textbf{\begin{CJK}{UTF8}{gbsn}开\end{CJK}} & \textbf{\begin{CJK}{UTF8}{min}開いた\end{CJK}} & \textbf{\begin{CJK}{UTF8}{mj}열린\end{CJK}} \\*
\textit{closed} & \textit{fermé} & \textit{geschlossen} & \textit{cerrado} & \textit{\begin{CJK}{UTF8}{gbsn}关\end{CJK}} & \textit{\begin{CJK}{UTF8}{min}閉じた\end{CJK}} & \textit{\begin{CJK}{UTF8}{mj}닫힌\end{CJK}} \\ \midrule
\textbf{rich} & \textbf{riche} & \textbf{reich} & \textbf{rico} & \textbf{\begin{CJK}{UTF8}{gbsn}富裕\end{CJK}} & \textbf{\begin{CJK}{UTF8}{min}裕福な\end{CJK}} & \textbf{\begin{CJK}{UTF8}{mj}부유한\end{CJK}} \\*
\textit{poor} & \textit{pauvre} & \textit{arm} & \textit{pobre} & \textit{\begin{CJK}{UTF8}{gbsn}贫穷\end{CJK}} & \textit{\begin{CJK}{UTF8}{min}貧しい\end{CJK}} & \textit{\begin{CJK}{UTF8}{mj}가난한\end{CJK}} \\ \midrule
\textbf{beautiful} & \textbf{beau} & \textbf{schön} & \textbf{hermoso} & \textbf{\begin{CJK}{UTF8}{gbsn}美\end{CJK}} & \textbf{\begin{CJK}{UTF8}{min}美しい\end{CJK}} & \textbf{\begin{CJK}{UTF8}{mj}아름다운\end{CJK}} \\*
\textit{ugly} & \textit{laid} & \textit{hässlich} & \textit{feo} & \textit{\begin{CJK}{UTF8}{gbsn}丑\end{CJK}} & \textit{\begin{CJK}{UTF8}{min}醜い\end{CJK}} & \textit{\begin{CJK}{UTF8}{mj}못생긴\end{CJK}} \\ \midrule
\textbf{long} & \textbf{long} & \textbf{lang} & \textbf{largo} & \textbf{\begin{CJK}{UTF8}{gbsn}长\end{CJK}} & \textbf{\begin{CJK}{UTF8}{min}長い\end{CJK}} & \textbf{\begin{CJK}{UTF8}{mj}긴\end{CJK}} \\*
\textit{short} & \textit{court} & \textit{kurz} & \textit{corto} & \textit{\begin{CJK}{UTF8}{gbsn}短\end{CJK}} & \textit{\begin{CJK}{UTF8}{min}短い\end{CJK}} & \textit{\begin{CJK}{UTF8}{mj}짧은\end{CJK}} \\ \midrule
\textbf{wide} & \textbf{large} & \textbf{breit} & \textbf{ancho} & \textbf{\begin{CJK}{UTF8}{gbsn}宽\end{CJK}} & \textbf{\begin{CJK}{UTF8}{min}広い\end{CJK}} & \textbf{\begin{CJK}{UTF8}{mj}넓은\end{CJK}} \\*
\textit{narrow} & \textit{étroit} & \textit{eng} & \textit{estrecho} & \textit{\begin{CJK}{UTF8}{gbsn}窄\end{CJK}} & \textit{\begin{CJK}{UTF8}{min}狭い\end{CJK}} & \textit{\begin{CJK}{UTF8}{mj}좁은\end{CJK}} \\ \midrule
\textbf{hard} & \textbf{dur} & \textbf{hart} & \textbf{duro} & \textbf{\begin{CJK}{UTF8}{gbsn}硬\end{CJK}} & \textbf{\begin{CJK}{UTF8}{min}硬い\end{CJK}} & \textbf{\begin{CJK}{UTF8}{mj}딱딱한\end{CJK}} \\*
\textit{soft} & \textit{mou} & \textit{weich} & \textit{suave} & \textit{\begin{CJK}{UTF8}{gbsn}软\end{CJK}} & \textit{\begin{CJK}{UTF8}{min}柔らかい\end{CJK}} & \textit{\begin{CJK}{UTF8}{mj}부드러운\end{CJK}} \\ \midrule
\textbf{dry} & \textbf{sec} & \textbf{trocken} & \textbf{seco} & \textbf{\begin{CJK}{UTF8}{gbsn}干\end{CJK}} & \textbf{\begin{CJK}{UTF8}{min}乾いた\end{CJK}} & \textbf{\begin{CJK}{UTF8}{mj}마른\end{CJK}} \\*
\textit{wet, moist} & \textit{humide, mouillé} & \textit{nass} & \textit{mojado, húmedo} & \textit{\begin{CJK}{UTF8}{gbsn}湿\end{CJK}} & \textit{\begin{CJK}{UTF8}{min}濡れた, 湿った\end{CJK}} & \textit{\begin{CJK}{UTF8}{mj}젖은, 축축한\end{CJK}} \\ \midrule
\textbf{loud} & \textbf{fort} & \textbf{laut} & \textbf{ruidoso} & \textbf{\begin{CJK}{UTF8}{gbsn}大声\end{CJK}} & \textbf{\begin{CJK}{UTF8}{min}うるさい\end{CJK}} & \textbf{\begin{CJK}{UTF8}{mj}시끄러운\end{CJK}} \\*
\textit{quiet, silent} & \textit{silencieux, doux} & \textit{leise, still} & \textit{tranquilo, silencioso} & \textit{\begin{CJK}{UTF8}{gbsn}安静, 小声\end{CJK}} & \textit{\begin{CJK}{UTF8}{min}静かな\end{CJK}} & \textit{\begin{CJK}{UTF8}{mj}조용한, 고요한\end{CJK}} \\ \midrule
\textbf{early} & \textbf{tôt} & \textbf{früh} & \textbf{temprano} & \textbf{\begin{CJK}{UTF8}{gbsn}早\end{CJK}} & \textbf{\begin{CJK}{UTF8}{min}早い\end{CJK}} & \textbf{\begin{CJK}{UTF8}{mj}이른\end{CJK}} \\*
\textit{late} & \textit{tard} & \textit{spät} & \textit{tarde} & \textit{\begin{CJK}{UTF8}{gbsn}晚\end{CJK}} & \textit{\begin{CJK}{UTF8}{min}遅い\end{CJK}} & \textit{\begin{CJK}{UTF8}{mj}늦은\end{CJK}} \\ \midrule
\textbf{near} & \textbf{proche} & \textbf{nah} & \textbf{cerca} & \textbf{\begin{CJK}{UTF8}{gbsn}近\end{CJK}} & \textbf{\begin{CJK}{UTF8}{min}近い\end{CJK}} & \textbf{\begin{CJK}{UTF8}{mj}가까운\end{CJK}} \\*
\textit{far} & \textit{loin} & \textit{fern, weit} & \textit{lejos} & \textit{\begin{CJK}{UTF8}{gbsn}远\end{CJK}} & \textit{\begin{CJK}{UTF8}{min}遠い\end{CJK}} & \textit{\begin{CJK}{UTF8}{mj}먼\end{CJK}} \\ \midrule
\textbf{deep} & \textbf{profond} & \textbf{tief} & \textbf{profundo} & \textbf{\begin{CJK}{UTF8}{gbsn}深\end{CJK}} & \textbf{\begin{CJK}{UTF8}{min}深い\end{CJK}} & \textbf{\begin{CJK}{UTF8}{mj}깊은\end{CJK}} \\*
\textit{shallow} & \textit{peu profond} & \textit{flach} & \textit{superficial, poco profundo} & \textit{\begin{CJK}{UTF8}{gbsn}浅\end{CJK}} & \textit{\begin{CJK}{UTF8}{min}浅い\end{CJK}} & \textit{\begin{CJK}{UTF8}{mj}얕은\end{CJK}} \\ \midrule
\textbf{bad} & \textbf{mauvais} & \textbf{schlecht} & \textbf{malo} & \textbf{\begin{CJK}{UTF8}{gbsn}坏\end{CJK}} & \textbf{\begin{CJK}{UTF8}{min}悪い\end{CJK}} & \textbf{\begin{CJK}{UTF8}{mj}나쁜\end{CJK}} \\*
\textit{good} & \textit{bon} & \textit{gut} & \textit{bueno} & \textit{\begin{CJK}{UTF8}{gbsn}好\end{CJK}} & \textit{\begin{CJK}{UTF8}{min}良い\end{CJK}} & \textit{\begin{CJK}{UTF8}{mj}좋은\end{CJK}} \\ \midrule
\textbf{sad} & \textbf{triste} & \textbf{traurig} & \textbf{triste} & \textbf{\begin{CJK}{UTF8}{gbsn}难过\end{CJK}} & \textbf{\begin{CJK}{UTF8}{min}悲しい\end{CJK}} & \textbf{\begin{CJK}{UTF8}{mj}슬픈\end{CJK}} \\*
\textit{happy, joyful} & \textit{heureux, joyeux} & \textit{glücklich} & \textit{feliz, alegre} & \textit{\begin{CJK}{UTF8}{gbsn}开心\end{CJK}} & \textit{\begin{CJK}{UTF8}{min}嬉しい\end{CJK}} & \textit{\begin{CJK}{UTF8}{mj}행복한, 기쁜\end{CJK}} \\ \midrule
\textbf{small} & \textbf{petit} & \textbf{klein} & \textbf{pequeño} & \textbf{\begin{CJK}{UTF8}{gbsn}小\end{CJK}} & \textbf{\begin{CJK}{UTF8}{min}小さい\end{CJK}} & \textbf{\begin{CJK}{UTF8}{mj}작은\end{CJK}} \\*
\textit{big, large} & \textit{grand} & \textit{groß} & \textit{grande} & \textit{\begin{CJK}{UTF8}{gbsn}大\end{CJK}} & \textit{\begin{CJK}{UTF8}{min}大きい\end{CJK}} & \textit{\begin{CJK}{UTF8}{mj}큰\end{CJK}} \\ \midrule
\textbf{cold} & \textbf{froid} & \textbf{kalt} & \textbf{frío} & \textbf{\begin{CJK}{UTF8}{gbsn}冷\end{CJK}} & \textbf{\begin{CJK}{UTF8}{min}寒い\end{CJK}} & \textbf{\begin{CJK}{UTF8}{mj}추운\end{CJK}} \\*
\textit{hot, warm} & \textit{chaud} & \textit{heiß} & \textit{caliente, cálido} & \textit{\begin{CJK}{UTF8}{gbsn}热\end{CJK}} & \textit{\begin{CJK}{UTF8}{min}暑い, 熱い\end{CJK}} & \textit{\begin{CJK}{UTF8}{mj}더운, 따뜻한\end{CJK}} \\ \midrule
\textbf{slow} & \textbf{lent} & \textbf{langsam} & \textbf{lento} & \textbf{\begin{CJK}{UTF8}{gbsn}慢\end{CJK}} & \textbf{\begin{CJK}{UTF8}{min}遅い\end{CJK}} & \textbf{\begin{CJK}{UTF8}{mj}느린\end{CJK}} \\*
\textit{fast, quick} & \textit{rapide} & \textit{schnell} & \textit{rápido, veloz} & \textit{\begin{CJK}{UTF8}{gbsn}快\end{CJK}} & \textit{\begin{CJK}{UTF8}{min}速い, 早い\end{CJK}} & \textit{\begin{CJK}{UTF8}{mj}빠른\end{CJK}} \\ \midrule
\textbf{heavy} & \textbf{lourd} & \textbf{schwer} & \textbf{pesado} & \textbf{\begin{CJK}{UTF8}{gbsn}重\end{CJK}} & \textbf{\begin{CJK}{UTF8}{min}重い\end{CJK}} & \textbf{\begin{CJK}{UTF8}{mj}무거운\end{CJK}} \\*
\textit{light} & \textit{léger} & \textit{leicht} & \textit{ligero} & \textit{\begin{CJK}{UTF8}{gbsn}轻\end{CJK}} & \textit{\begin{CJK}{UTF8}{min}軽い\end{CJK}} & \textit{\begin{CJK}{UTF8}{mj}가벼운\end{CJK}} \\ \midrule
\textbf{difficult} & \textbf{difficile} & \textbf{schwierig} & \textbf{difícil} & \textbf{\begin{CJK}{UTF8}{gbsn}难\end{CJK}} & \textbf{\begin{CJK}{UTF8}{min}難しい\end{CJK}} & \textbf{\begin{CJK}{UTF8}{mj}어려운\end{CJK}} \\*
\textit{easy, simple} & \textit{facile} & \textit{einfach} & \textit{fácil, sencillo} & \textit{\begin{CJK}{UTF8}{gbsn}容易, 易\end{CJK}} & \textit{\begin{CJK}{UTF8}{min}簡単な, 簡単\end{CJK}} & \textit{\begin{CJK}{UTF8}{mj}쉬운, 간단한\end{CJK}} \\ \midrule
\textbf{old} & \textbf{vieux} & \textbf{alt} & \textbf{viejo} & \textbf{\begin{CJK}{UTF8}{gbsn}旧\end{CJK}} & \textbf{\begin{CJK}{UTF8}{min}古い\end{CJK}} & \textbf{\begin{CJK}{UTF8}{mj}오래된\end{CJK}} \\*
\textit{new} & \textit{nouveau, jeune} & \textit{neu} & \textit{nuevo, joven} & \textit{\begin{CJK}{UTF8}{gbsn}新\end{CJK}} & \textit{\begin{CJK}{UTF8}{min}新しい\end{CJK}} & \textit{\begin{CJK}{UTF8}{mj}새로운, 젊은\end{CJK}} \\ \midrule
\textbf{false} & \textbf{faux} & \textbf{falsch} & \textbf{falso} & \textbf{\begin{CJK}{UTF8}{gbsn}假\end{CJK}} & \textbf{\begin{CJK}{UTF8}{min}間違った\end{CJK}} & \textbf{\begin{CJK}{UTF8}{mj}거짓의\end{CJK}} \\*
\textit{true, correct} & \textit{vrai, correct} & \textit{wahr, richtig} & \textit{verdadero, correcto} & \textit{\begin{CJK}{UTF8}{gbsn}真\end{CJK}} & \textit{\begin{CJK}{UTF8}{min}正しい\end{CJK}} & \textit{\begin{CJK}{UTF8}{mj}진실한, 올바른\end{CJK}} \\ \midrule
\textbf{dead} & \textbf{mort} & \textbf{tot} & \textbf{muerto} & \textbf{\begin{CJK}{UTF8}{gbsn}死\end{CJK}} & \textbf{\begin{CJK}{UTF8}{min}死んでいる\end{CJK}} & \textbf{\begin{CJK}{UTF8}{mj}죽은\end{CJK}} \\*
\textit{alive, living} & \textit{vivant} & \textit{lebendig} & \textit{vivo, viviente} & \textit{\begin{CJK}{UTF8}{gbsn}活, 生\end{CJK}} & \textit{\begin{CJK}{UTF8}{min}生きている\end{CJK}} & \textit{\begin{CJK}{UTF8}{mj}살아있는, 생생한\end{CJK}} \\ \midrule
\textbf{empty} & \textbf{vide} & \textbf{leer} & \textbf{vacío} & \textbf{\begin{CJK}{UTF8}{gbsn}空\end{CJK}} & \textbf{\begin{CJK}{UTF8}{min}空っぽ\end{CJK}} & \textbf{\begin{CJK}{UTF8}{mj}빈\end{CJK}} \\*
\textit{full} & \textit{plein} & \textit{voll} & \textit{lleno} & \textit{\begin{CJK}{UTF8}{gbsn}满\end{CJK}} & \textit{\begin{CJK}{UTF8}{min}満杯\end{CJK}} & \textit{\begin{CJK}{UTF8}{mj}가득 찬\end{CJK}} \\ \midrule
\textbf{dark} & \textbf{sombre} & \textbf{dunkel} & \textbf{oscuro} & \textbf{\begin{CJK}{UTF8}{gbsn}暗\end{CJK}} & \textbf{\begin{CJK}{UTF8}{min}暗い\end{CJK}} & \textbf{\begin{CJK}{UTF8}{mj}어두운\end{CJK}} \\*
\textit{bright, light} & \textit{brillant, clair} & \textit{hell} & \textit{brillante, claro} & \textit{\begin{CJK}{UTF8}{gbsn}亮, 光明\end{CJK}} & \textit{\begin{CJK}{UTF8}{min}明るい\end{CJK}} & \textit{\begin{CJK}{UTF8}{mj}밝은\end{CJK}} \\ \midrule
\textbf{weak} & \textbf{faible} & \textbf{schwach} & \textbf{débil} & \textbf{\begin{CJK}{UTF8}{gbsn}弱\end{CJK}} & \textbf{\begin{CJK}{UTF8}{min}弱い\end{CJK}} & \textbf{\begin{CJK}{UTF8}{mj}약한\end{CJK}} \\*
\textit{strong} & \textit{fort} & \textit{stark} & \textit{fuerte} & \textit{\begin{CJK}{UTF8}{gbsn}强, 强壮\end{CJK}} & \textit{\begin{CJK}{UTF8}{min}強い, 丈夫な\end{CJK}} & \textit{\begin{CJK}{UTF8}{mj}강한\end{CJK}} \\ \midrule
\textbf{dirty} & \textbf{sale} & \textbf{schmutzig} & \textbf{sucio} & \textbf{\begin{CJK}{UTF8}{gbsn}脏\end{CJK}} & \textbf{\begin{CJK}{UTF8}{min}汚い\end{CJK}} & \textbf{\begin{CJK}{UTF8}{mj}더러운\end{CJK}} \\*
\textit{clean} & \textit{propre} & \textit{sauber} & \textit{limpio} & \textit{\begin{CJK}{UTF8}{gbsn}干净\end{CJK}} & \textit{\begin{CJK}{UTF8}{min}きれいな, きれい, 綺麗\end{CJK}} & \textit{\begin{CJK}{UTF8}{mj}깨끗한\end{CJK}} \\ \midrule
\textbf{closed} & \textbf{fermé} & \textbf{geschlossen} & \textbf{cerrado} & \textbf{\begin{CJK}{UTF8}{gbsn}关\end{CJK}} & \textbf{\begin{CJK}{UTF8}{min}閉じた\end{CJK}} & \textbf{\begin{CJK}{UTF8}{mj}닫힌\end{CJK}} \\*
\textit{open} & \textit{ouvert} & \textit{offen} & \textit{abierto} & \textit{\begin{CJK}{UTF8}{gbsn}开\end{CJK}} & \textit{\begin{CJK}{UTF8}{min}開いた\end{CJK}} & \textit{\begin{CJK}{UTF8}{mj}열린\end{CJK}} \\ \midrule
\textbf{poor} & \textbf{pauvre} & \textbf{arm} & \textbf{pobre} & \textbf{\begin{CJK}{UTF8}{gbsn}贫穷\end{CJK}} & \textbf{\begin{CJK}{UTF8}{min}貧しい\end{CJK}} & \textbf{\begin{CJK}{UTF8}{mj}가난한\end{CJK}} \\*
\textit{rich, wealthy} & \textit{riche} & \textit{reich} & \textit{rico, adinerado} & \textit{\begin{CJK}{UTF8}{gbsn}富裕\end{CJK}} & \textit{\begin{CJK}{UTF8}{min}裕福な, 豊かな\end{CJK}} & \textit{\begin{CJK}{UTF8}{mj}부유한, 풍부한\end{CJK}} \\ \midrule
\textbf{ugly} & \textbf{laid} & \textbf{hässlich} & \textbf{feo} & \textbf{\begin{CJK}{UTF8}{gbsn}丑\end{CJK}} & \textbf{\begin{CJK}{UTF8}{min}醜い\end{CJK}} & \textbf{\begin{CJK}{UTF8}{mj}못생긴\end{CJK}} \\*
\textit{beautiful, pretty} & \textit{beau, joli} & \textit{schön} & \textit{hermoso, bonito} & \textit{\begin{CJK}{UTF8}{gbsn}美\end{CJK}} & \textit{\begin{CJK}{UTF8}{min}美しい\end{CJK}} & \textit{\begin{CJK}{UTF8}{mj}아름다운, 예쁜\end{CJK}} \\ \midrule
\textbf{short} & \textbf{court} & \textbf{kurz} & \textbf{corto} & \textbf{\begin{CJK}{UTF8}{gbsn}短\end{CJK}} & \textbf{\begin{CJK}{UTF8}{min}短い\end{CJK}} & \textbf{\begin{CJK}{UTF8}{mj}짧은\end{CJK}} \\*
\textit{long} & \textit{long} & \textit{lang} & \textit{largo} & \textit{\begin{CJK}{UTF8}{gbsn}长\end{CJK}} & \textit{\begin{CJK}{UTF8}{min}長い\end{CJK}} & \textit{\begin{CJK}{UTF8}{mj}긴\end{CJK}} \\ \midrule
\textbf{narrow} & \textbf{étroit} & \textbf{eng} & \textbf{estrecho} & \textbf{\begin{CJK}{UTF8}{gbsn}窄\end{CJK}} & \textbf{\begin{CJK}{UTF8}{min}狭い\end{CJK}} & \textbf{\begin{CJK}{UTF8}{mj}좁은\end{CJK}} \\*
\textit{wide, broad} & \textit{large} & \textit{breit} & \textit{ancho} & \textit{\begin{CJK}{UTF8}{gbsn}宽\end{CJK}} & \textit{\begin{CJK}{UTF8}{min}広い\end{CJK}} & \textit{\begin{CJK}{UTF8}{mj}넓은\end{CJK}} \\ \midrule
\textbf{soft} & \textbf{mou} & \textbf{weich} & \textbf{suave} & \textbf{\begin{CJK}{UTF8}{gbsn}软\end{CJK}} & \textbf{\begin{CJK}{UTF8}{min}柔らかい\end{CJK}} & \textbf{\begin{CJK}{UTF8}{mj}부드러운\end{CJK}} \\*
\textit{hard, firm} & \textit{dur} & \textit{hart} & \textit{duro} & \textit{\begin{CJK}{UTF8}{gbsn}硬\end{CJK}} & \textit{\begin{CJK}{UTF8}{min}硬い\end{CJK}} & \textit{\begin{CJK}{UTF8}{mj}딱딱한\end{CJK}} \\ \midrule
\textbf{wet} & \textbf{humide} & \textbf{nass} & \textbf{mojado} & \textbf{\begin{CJK}{UTF8}{gbsn}湿\end{CJK}} & \textbf{\begin{CJK}{UTF8}{min}濡れた\end{CJK}} & \textbf{\begin{CJK}{UTF8}{mj}젖은\end{CJK}} \\*
\textit{dry} & \textit{sec} & \textit{trocken} & \textit{seco} & \textit{\begin{CJK}{UTF8}{gbsn}干\end{CJK}} & \textit{\begin{CJK}{UTF8}{min}乾いた\end{CJK}} & \textit{\begin{CJK}{UTF8}{mj}마른\end{CJK}} \\ \midrule
\textbf{quiet} & \textbf{silencieux} & \textbf{leise} & \textbf{tranquilo} & \textbf{\begin{CJK}{UTF8}{gbsn}安静\end{CJK}} & \textbf{\begin{CJK}{UTF8}{min}静かな\end{CJK}} & \textbf{\begin{CJK}{UTF8}{mj}조용한\end{CJK}} \\*
\textit{loud, noisy} & \textit{fort, bruyant} & \textit{laut} & \textit{ruidoso} & \textit{\begin{CJK}{UTF8}{gbsn}大声, 吵闹\end{CJK}} & \textit{\begin{CJK}{UTF8}{min}うるさい\end{CJK}} & \textit{\begin{CJK}{UTF8}{mj}시끄러운\end{CJK}} \\ \midrule
\textbf{late} & \textbf{tard} & \textbf{spät} & \textbf{tarde} & \textbf{\begin{CJK}{UTF8}{gbsn}晚\end{CJK}} & \textbf{\begin{CJK}{UTF8}{min}遅い\end{CJK}} & \textbf{\begin{CJK}{UTF8}{mj}늦은\end{CJK}} \\*
\textit{early} & \textit{tôt} & \textit{früh} & \textit{temprano} & \textit{\begin{CJK}{UTF8}{gbsn}早\end{CJK}} & \textit{\begin{CJK}{UTF8}{min}早い\end{CJK}} & \textit{\begin{CJK}{UTF8}{mj}이른\end{CJK}} \\ \midrule
\textbf{far} & \textbf{loin} & \textbf{fern} & \textbf{lejos} & \textbf{\begin{CJK}{UTF8}{gbsn}远\end{CJK}} & \textbf{\begin{CJK}{UTF8}{min}遠い\end{CJK}} & \textbf{\begin{CJK}{UTF8}{mj}먼\end{CJK}} \\*
\textit{near, close} & \textit{proche} & \textit{nah} & \textit{cerca} & \textit{\begin{CJK}{UTF8}{gbsn}近\end{CJK}} & \textit{\begin{CJK}{UTF8}{min}近い\end{CJK}} & \textit{\begin{CJK}{UTF8}{mj}가까운\end{CJK}} \\ \midrule
\textbf{shallow} & \textbf{peu profond} & \textbf{flach} & \textbf{superficial} & \textbf{\begin{CJK}{UTF8}{gbsn}浅\end{CJK}} & \textbf{\begin{CJK}{UTF8}{min}浅い\end{CJK}} & \textbf{\begin{CJK}{UTF8}{mj}얕은\end{CJK}} \\*
\textit{deep} & \textit{profond} & \textit{tief, hoch} & \textit{profundo} & \textit{\begin{CJK}{UTF8}{gbsn}深\end{CJK}} & \textit{\begin{CJK}{UTF8}{min}深い\end{CJK}} & \textit{\begin{CJK}{UTF8}{mj}깊은\end{CJK}} \\ \midrule
\textbf{young} & \textbf{jeune} & \textbf{jung} & \textbf{joven} & \textbf{\begin{CJK}{UTF8}{gbsn}年轻\end{CJK}} & \textbf{\begin{CJK}{UTF8}{min}若い\end{CJK}} & \textbf{\begin{CJK}{UTF8}{mj}젊은\end{CJK}} \\*
\textit{old} & \textit{vieux} & \textit{alt} & \textit{viejo} & \textit{\begin{CJK}{UTF8}{gbsn}老\end{CJK}} & \textit{\begin{CJK}{UTF8}{min}老いた\end{CJK}} & \textit{\begin{CJK}{UTF8}{mj}늙은\end{CJK}} \\ \midrule
\textbf{kind} & \textbf{gentil} & \textbf{freundlich} & \textbf{amable} & \textbf{\begin{CJK}{UTF8}{gbsn}善良\end{CJK}} & \textbf{\begin{CJK}{UTF8}{min}親切な\end{CJK}} & \textbf{\begin{CJK}{UTF8}{mj}친절한\end{CJK}} \\*
\textit{unkind, mean} & \textit{méchant} & \textit{unfreundlich} & \textit{desagradable, malo} & \textit{\begin{CJK}{UTF8}{gbsn}不善良, 邪恶\end{CJK}} & \textit{\begin{CJK}{UTF8}{min}意地悪な, 不親切な, 失礼な\end{CJK}} & \textit{\begin{CJK}{UTF8}{mj}불친절한, 못된\end{CJK}} \\ \midrule
\textbf{brave} & \textbf{courageux} & \textbf{mutig} & \textbf{valiente} & \textbf{\begin{CJK}{UTF8}{gbsn}勇敢\end{CJK}} & \textbf{\begin{CJK}{UTF8}{min}勇敢な\end{CJK}} & \textbf{\begin{CJK}{UTF8}{mj}용감한\end{CJK}} \\*
\textit{cowardly} & \textit{lâche} & \textit{feige} & \textit{cobarde} & \textit{\begin{CJK}{UTF8}{gbsn}懦弱\end{CJK}} & \textit{\begin{CJK}{UTF8}{min}臆病な\end{CJK}} & \textit{\begin{CJK}{UTF8}{mj}겁많은\end{CJK}} \\ \midrule
\textbf{wise} & \textbf{sage} & \textbf{weise} & \textbf{sabio} & \textbf{\begin{CJK}{UTF8}{gbsn}明智\end{CJK}} & \textbf{\begin{CJK}{UTF8}{min}賢い\end{CJK}} & \textbf{\begin{CJK}{UTF8}{mj}현명한\end{CJK}} \\*
\textit{foolish} & \textit{insensé, stupide} & \textit{dumm, töricht} & \textit{tonto, necio} & \textit{\begin{CJK}{UTF8}{gbsn}愚蠢\end{CJK}} & \textit{\begin{CJK}{UTF8}{min}愚かな\end{CJK}} & \textit{\begin{CJK}{UTF8}{mj}어리석은\end{CJK}} \\ \midrule
\textbf{polite} & \textbf{poli} & \textbf{höflich} & \textbf{educado} & \textbf{\begin{CJK}{UTF8}{gbsn}礼貌\end{CJK}} & \textbf{\begin{CJK}{UTF8}{min}丁寧な\end{CJK}} & \textbf{\begin{CJK}{UTF8}{mj}예의 바른\end{CJK}} \\*
\textit{impolite, rude} & \textit{impoli, grossier} & \textit{unhöflich, grob} & \textit{descortés, maleducado} & \textit{\begin{CJK}{UTF8}{gbsn}不礼貌\end{CJK}} & \textit{\begin{CJK}{UTF8}{min}失礼な, 無礼な, 粗略な, 粗雑な\end{CJK}} & \textit{\begin{CJK}{UTF8}{mj}무례한\end{CJK}} \\ \midrule
\textbf{patient} & \textbf{patient} & \textbf{geduldig} & \textbf{paciente} & \textbf{\begin{CJK}{UTF8}{gbsn}耐心\end{CJK}} & \textbf{\begin{CJK}{UTF8}{min}我慢強い\end{CJK}} & \textbf{\begin{CJK}{UTF8}{mj}인내심 있는\end{CJK}} \\*
\textit{impatient} & \textit{impatient} & \textit{ungeduldig} & \textit{impaciente} & \textit{\begin{CJK}{UTF8}{gbsn}不耐烦\end{CJK}} & \textit{\begin{CJK}{UTF8}{min}せっかちな\end{CJK}} & \textit{\begin{CJK}{UTF8}{mj}성급한, 참을성 없는\end{CJK}} \\ \midrule
\textbf{honest} & \textbf{honnête} & \textbf{ehrlich} & \textbf{honesto} & \textbf{\begin{CJK}{UTF8}{gbsn}诚实\end{CJK}} & \textbf{\begin{CJK}{UTF8}{min}正直な\end{CJK}} & \textbf{\begin{CJK}{UTF8}{mj}정직한\end{CJK}} \\*
\textit{dishonest} & \textit{malhonnête} & \textit{unehrlich} & \textit{deshonesto} & \textit{\begin{CJK}{UTF8}{gbsn}不诚实\end{CJK}} & \textit{\begin{CJK}{UTF8}{min}不正直な\end{CJK}} & \textit{\begin{CJK}{UTF8}{mj}부정직한\end{CJK}} \\ \midrule
\textbf{safe} & \textbf{sûr} & \textbf{sicher} & \textbf{seguro} & \textbf{\begin{CJK}{UTF8}{gbsn}安全\end{CJK}} & \textbf{\begin{CJK}{UTF8}{min}安全な\end{CJK}} & \textbf{\begin{CJK}{UTF8}{mj}안전한\end{CJK}} \\*
\textit{dangerous, unsafe} & \textit{dangereux} & \textit{gefährlich, unsicher} & \textit{peligroso, inseguro} & \textit{\begin{CJK}{UTF8}{gbsn}危险\end{CJK}} & \textit{\begin{CJK}{UTF8}{min}危険な\end{CJK}} & \textit{\begin{CJK}{UTF8}{mj}위험한\end{CJK}} \\ \midrule
\textbf{active} & \textbf{actif} & \textbf{aktiv} & \textbf{activo} & \textbf{\begin{CJK}{UTF8}{gbsn}积极\end{CJK}} & \textbf{\begin{CJK}{UTF8}{min}活動的な\end{CJK}} & \textbf{\begin{CJK}{UTF8}{mj}활동적인\end{CJK}} \\*
\textit{inactive, passive} & \textit{inactif, passif} & \textit{inaktiv, passiv} & \textit{inactivo, pasivo} & \textit{\begin{CJK}{UTF8}{gbsn}消极\end{CJK}} & \textit{\begin{CJK}{UTF8}{min}消極的な\end{CJK}} & \textit{\begin{CJK}{UTF8}{mj}비활동적인, 소극적인\end{CJK}} \\ \midrule
\textbf{straight} & \textbf{droit} & \textbf{gerade} & \textbf{recto} & \textbf{\begin{CJK}{UTF8}{gbsn}直\end{CJK}} & \textbf{\begin{CJK}{UTF8}{min}まっすぐな\end{CJK}} & \textbf{\begin{CJK}{UTF8}{mj}곧은\end{CJK}} \\*
\textit{curved, bent} & \textit{courbe, tordu} & \textit{gebogen, krumm} & \textit{curvo, doblado} & \textit{\begin{CJK}{UTF8}{gbsn}弯\end{CJK}} & \textit{\begin{CJK}{UTF8}{min}曲がった\end{CJK}} & \textit{\begin{CJK}{UTF8}{mj}굽은, 휘어진\end{CJK}} \\ \midrule
\textbf{whole} & \textbf{entier} & \textbf{ganz} & \textbf{entero} & \textbf{\begin{CJK}{UTF8}{gbsn}完整\end{CJK}} & \textbf{\begin{CJK}{UTF8}{min}全体の\end{CJK}} & \textbf{\begin{CJK}{UTF8}{mj}전체의\end{CJK}} \\*
\textit{part, broken} & \textit{partiel, cassé} & \textit{teilweise, gebrochen} & \textit{parcial, roto} & \textit{\begin{CJK}{UTF8}{gbsn}部分, 破\end{CJK}} & \textit{\begin{CJK}{UTF8}{min}部分的な, 壊れた\end{CJK}} & \textit{\begin{CJK}{UTF8}{mj}부분적인, 부서진\end{CJK}} \\ \midrule
\textbf{calm} & \textbf{calme} & \textbf{ruhig} & \textbf{calmado} & \textbf{\begin{CJK}{UTF8}{gbsn}平静\end{CJK}} & \textbf{\begin{CJK}{UTF8}{min}穏やかな\end{CJK}} & \textbf{\begin{CJK}{UTF8}{mj}차분한\end{CJK}} \\*
\textit{agitated, stormy} & \textit{agité, orageux} & \textit{aufgeregt, stürmisch} & \textit{agitado, tormentoso} & \textit{\begin{CJK}{UTF8}{gbsn}激动, 骚乱\end{CJK}} & \textit{\begin{CJK}{UTF8}{min}荒れた, 興奮した\end{CJK}} & \textit{\begin{CJK}{UTF8}{mj}격앙된, 폭풍우치는\end{CJK}} \\ \midrule
\textbf{correct} & \textbf{correct} & \textbf{richtig} & \textbf{correcto} & \textbf{\begin{CJK}{UTF8}{gbsn}正确\end{CJK}} & \textbf{\begin{CJK}{UTF8}{min}正しい\end{CJK}} & \textbf{\begin{CJK}{UTF8}{mj}정확한\end{CJK}} \\*
\textit{incorrect, wrong} & \textit{incorrect, faux} & \textit{falsch} & \textit{incorrecto, equivocado} & \textit{\begin{CJK}{UTF8}{gbsn}不正确, 错误\end{CJK}} & \textit{\begin{CJK}{UTF8}{min}間違っている\end{CJK}} & \textit{\begin{CJK}{UTF8}{mj}부정확한, 틀린\end{CJK}} \\ \midrule
\textbf{complex} & \textbf{complexe} & \textbf{komplex} & \textbf{complejo} & \textbf{\begin{CJK}{UTF8}{gbsn}复杂\end{CJK}} & \textbf{\begin{CJK}{UTF8}{min}複雑な\end{CJK}} & \textbf{\begin{CJK}{UTF8}{mj}복잡한\end{CJK}} \\*
\textit{simple} & \textit{simple} & \textit{einfach} & \textit{simple, sencillo} & \textit{\begin{CJK}{UTF8}{gbsn}简单\end{CJK}} & \textit{\begin{CJK}{UTF8}{min}単純な\end{CJK}} & \textit{\begin{CJK}{UTF8}{mj}간단한, 단순한\end{CJK}} \\ \midrule
\textbf{effective} & \textbf{efficace} & \textbf{effektiv} & \textbf{efectivo} & \textbf{\begin{CJK}{UTF8}{gbsn}有效\end{CJK}} & \textbf{\begin{CJK}{UTF8}{min}効果的な\end{CJK}} & \textbf{\begin{CJK}{UTF8}{mj}효과적인\end{CJK}} \\*
\textit{ineffective} & \textit{inefficace} & \textit{ineffektiv} & \textit{ineficaz} & \textit{\begin{CJK}{UTF8}{gbsn}无效\end{CJK}} & \textit{\begin{CJK}{UTF8}{min}非効果的な\end{CJK}} & \textit{\begin{CJK}{UTF8}{mj}비효과적인\end{CJK}} \\ \midrule
\textbf{famous} & \textbf{célèbre} & \textbf{berühmt} & \textbf{famoso} & \textbf{\begin{CJK}{UTF8}{gbsn}著名\end{CJK}} & \textbf{\begin{CJK}{UTF8}{min}有名な\end{CJK}} & \textbf{\begin{CJK}{UTF8}{mj}유명한\end{CJK}} \\*
\textit{unknown, obscure} & \textit{inconnu, obscur} & \textit{unbekannt, unbedeutend} & \textit{desconocido, oscuro} & \textit{\begin{CJK}{UTF8}{gbsn}无名, 不为人知\end{CJK}} & \textit{\begin{CJK}{UTF8}{min}無名の, 知られていない\end{CJK}} & \textit{\begin{CJK}{UTF8}{mj}무명의, 잘 알려지지 않은\end{CJK}} \\ \midrule
\textbf{generous} & \textbf{généreux} & \textbf{großzügig} & \textbf{generoso} & \textbf{\begin{CJK}{UTF8}{gbsn}慷慨\end{CJK}} & \textbf{\begin{CJK}{UTF8}{min}寛大な\end{CJK}} & \textbf{\begin{CJK}{UTF8}{mj}관대한\end{CJK}} \\*
\textit{stingy, mean} & \textit{avare, mesquin} & \textit{geizig} & \textit{tacaño, malo} & \textit{\begin{CJK}{UTF8}{gbsn}吝啬\end{CJK}} & \textit{\begin{CJK}{UTF8}{min}ケチな\end{CJK}} & \textit{\begin{CJK}{UTF8}{mj}인색한, 못된\end{CJK}} \\ \midrule
\textbf{content} & \textbf{content} & \textbf{zufrieden} & \textbf{contento} & \textbf{\begin{CJK}{UTF8}{gbsn}高兴\end{CJK}} & \textbf{\begin{CJK}{UTF8}{min}幸せな\end{CJK}} & \textbf{\begin{CJK}{UTF8}{mj}행복한\end{CJK}} \\*
\textit{unhappy, dissatisfied} & \textit{mécontent, insatisfait} & \textit{unzufrieden} & \textit{infeliz, insatisfecho} & \textit{\begin{CJK}{UTF8}{gbsn}不高兴, 不满意\end{CJK}} & \textit{\begin{CJK}{UTF8}{min}不幸せな\end{CJK}} & \textit{\begin{CJK}{UTF8}{mj}불행한, 불만족스러운\end{CJK}} \\ \midrule
\textbf{healthy} & \textbf{sain} & \textbf{gesund} & \textbf{saludable} & \textbf{\begin{CJK}{UTF8}{gbsn}健康\end{CJK}} & \textbf{\begin{CJK}{UTF8}{min}健康な\end{CJK}} & \textbf{\begin{CJK}{UTF8}{mj}건강한\end{CJK}} \\*
\textit{unhealthy, sick} & \textit{malsain, malade} & \textit{ungesund, krank} & \textit{insalubre, enfermo} & \textit{\begin{CJK}{UTF8}{gbsn}不健康, 生病\end{CJK}} & \textit{\begin{CJK}{UTF8}{min}不健康な, 病気の\end{CJK}} & \textit{\begin{CJK}{UTF8}{mj}건강하지 않은, 아픈\end{CJK}} \\ \midrule
\textbf{high} & \textbf{haut} & \textbf{hoch} & \textbf{alto} & \textbf{\begin{CJK}{UTF8}{gbsn}高\end{CJK}} & \textbf{\begin{CJK}{UTF8}{min}高い\end{CJK}} & \textbf{\begin{CJK}{UTF8}{mj}높은\end{CJK}} \\*
\textit{low} & \textit{bas} & \textit{niedrig} & \textit{bajo} & \textit{\begin{CJK}{UTF8}{gbsn}低\end{CJK}} & \textit{\begin{CJK}{UTF8}{min}低い\end{CJK}} & \textit{\begin{CJK}{UTF8}{mj}낮은\end{CJK}} \\ \midrule
\textbf{important} & \textbf{important} & \textbf{wichtig} & \textbf{importante} & \textbf{\begin{CJK}{UTF8}{gbsn}重要\end{CJK}} & \textbf{\begin{CJK}{UTF8}{min}重要な\end{CJK}} & \textbf{\begin{CJK}{UTF8}{mj}중요한\end{CJK}} \\*
\textit{unimportant, trivial} & \textit{sans importance, insignifiant} & \textit{unwichtig, trivial} & \textit{sin importance, trivial} & \textit{\begin{CJK}{UTF8}{gbsn}不重要, 琐碎\end{CJK}} & \textit{\begin{CJK}{UTF8}{min}重要でない, 取るに足らない\end{CJK}} & \textit{\begin{CJK}{UTF8}{mj}중요하지 않은, 사소한\end{CJK}} \\ \midrule
\textbf{innocent} & \textbf{innocent} & \textbf{unschuldig} & \textbf{inocente} & \textbf{\begin{CJK}{UTF8}{gbsn}无辜\end{CJK}} & \textbf{\begin{CJK}{UTF8}{min}無罪の\end{CJK}} & \textbf{\begin{CJK}{UTF8}{mj}무고한\end{CJK}} \\*
\textit{guilty} & \textit{coupable} & \textit{schuldig} & \textit{culpable} & \textit{\begin{CJK}{UTF8}{gbsn}有罪\end{CJK}} & \textit{\begin{CJK}{UTF8}{min}有罪の\end{CJK}} & \textit{\begin{CJK}{UTF8}{mj}유죄의\end{CJK}} \\ \midrule
\textbf{known} & \textbf{connu} & \textbf{bekannt} & \textbf{conocido} & \textbf{\begin{CJK}{UTF8}{gbsn}已知\end{CJK}} & \textbf{\begin{CJK}{UTF8}{min}既知の\end{CJK}} & \textbf{\begin{CJK}{UTF8}{mj}알려진\end{CJK}} \\*
\textit{unknown} & \textit{inconnu} & \textit{unbekannt} & \textit{desconocido} & \textit{\begin{CJK}{UTF8}{gbsn}未知\end{CJK}} & \textit{\begin{CJK}{UTF8}{min}未知の\end{CJK}} & \textit{\begin{CJK}{UTF8}{mj}알려지지 않은\end{CJK}} \\ \midrule
\textbf{male} & \textbf{masculin} & \textbf{männlich} & \textbf{masculino} & \textbf{\begin{CJK}{UTF8}{gbsn}男性\end{CJK}} & \textbf{\begin{CJK}{UTF8}{min}男性の\end{CJK}} & \textbf{\begin{CJK}{UTF8}{mj}남성의\end{CJK}} \\*
\textit{female} & \textit{féminin} & \textit{weiblich} & \textit{femenino} & \textit{\begin{CJK}{UTF8}{gbsn}女性\end{CJK}} & \textit{\begin{CJK}{UTF8}{min}女性の\end{CJK}} & \textit{\begin{CJK}{UTF8}{mj}여성의\end{CJK}} \\ \midrule
\textbf{normal} & \textbf{normal} & \textbf{normal} & \textbf{normal} & \textbf{\begin{CJK}{UTF8}{gbsn}正常\end{CJK}} & \textbf{\begin{CJK}{UTF8}{min}普通の\end{CJK}} & \textbf{\begin{CJK}{UTF8}{mj}정상적인\end{CJK}} \\*
\textit{abnormal, unusual} & \textit{anormal, inusuel} & \textit{abnormal, ungewöhnlich} & \textit{anormal, inusual} & \textit{\begin{CJK}{UTF8}{gbsn}异常, 不正常\end{CJK}} & \textit{\begin{CJK}{UTF8}{min}異常な\end{CJK}} & \textit{\begin{CJK}{UTF8}{mj}비정상적인, 이상한\end{CJK}} \\ \midrule
\textbf{possible} & \textbf{possible} & \textbf{möglich} & \textbf{posible} & \textbf{\begin{CJK}{UTF8}{gbsn}可能\end{CJK}} & \textbf{\begin{CJK}{UTF8}{min}可能な\end{CJK}} & \textbf{\begin{CJK}{UTF8}{mj}가능한\end{CJK}} \\*
\textit{impossible} & \textit{impossible} & \textit{unmöglich} & \textit{imposible} & \textit{\begin{CJK}{UTF8}{gbsn}不可能\end{CJK}} & \textit{\begin{CJK}{UTF8}{min}不可能な\end{CJK}} & \textit{\begin{CJK}{UTF8}{mj}불가능한\end{CJK}} \\ \midrule
\textbf{private} & \textbf{privé} & \textbf{privat} & \textbf{privado} & \textbf{\begin{CJK}{UTF8}{gbsn}私人\end{CJK}} & \textbf{\begin{CJK}{UTF8}{min}個人の\end{CJK}} & \textbf{\begin{CJK}{UTF8}{mj}사적인\end{CJK}} \\*
\textit{public} & \textit{public} & \textit{öffentlich} & \textit{público} & \textit{\begin{CJK}{UTF8}{gbsn}公共\end{CJK}} & \textit{\begin{CJK}{UTF8}{min}公共の\end{CJK}} & \textit{\begin{CJK}{UTF8}{mj}공적인\end{CJK}} \\ \midrule
\textbf{right} & \textbf{juste} & \textbf{richtig} & \textbf{correcto} & \textbf{\begin{CJK}{UTF8}{gbsn}对\end{CJK}} & \textbf{\begin{CJK}{UTF8}{min}正しい\end{CJK}} & \textbf{\begin{CJK}{UTF8}{mj}올바른\end{CJK}} \\*
\textit{wrong} & \textit{faux} & \textit{falsch} & \textit{incorrecto, equivocado} & \textit{\begin{CJK}{UTF8}{gbsn}错\end{CJK}} & \textit{\begin{CJK}{UTF8}{min}間違っている\end{CJK}} & \textit{\begin{CJK}{UTF8}{mj}틀린\end{CJK}} \\ \midrule
\textbf{simple} & \textbf{simple} & \textbf{einfach} & \textbf{sencillo} & \textbf{\begin{CJK}{UTF8}{gbsn}简单\end{CJK}} & \textbf{\begin{CJK}{UTF8}{min}簡単な\end{CJK}} & \textbf{\begin{CJK}{UTF8}{mj}단순한\end{CJK}} \\*
\textit{complex} & \textit{complexe} & \textit{komplex} & \textit{complejo} & \textit{\begin{CJK}{UTF8}{gbsn}复杂\end{CJK}} & \textit{\begin{CJK}{UTF8}{min}複雑な\end{CJK}} & \textit{\begin{CJK}{UTF8}{mj}복잡한\end{CJK}} \\ \midrule
\textbf{sweet} & \textbf{doux} & \textbf{süß} & \textbf{dulce} & \textbf{\begin{CJK}{UTF8}{gbsn}甜\end{CJK}} & \textbf{\begin{CJK}{UTF8}{min}甘い\end{CJK}} & \textbf{\begin{CJK}{UTF8}{mj}달콤한\end{CJK}} \\*
\textit{sour, bitter} & \textit{acide, amer} & \textit{sauer, bitter} & \textit{agrio, amargo} & \textit{\begin{CJK}{UTF8}{gbsn}酸, 苦\end{CJK}} & \textit{\begin{CJK}{UTF8}{min}酸っぱい, 苦い\end{CJK}} & \textit{\begin{CJK}{UTF8}{mj}신, 쓴\end{CJK}} \\ \midrule
\textbf{visible} & \textbf{visible} & \textbf{sichtbar} & \textbf{visible} & \textbf{\begin{CJK}{UTF8}{gbsn}可见\end{CJK}} & \textbf{\begin{CJK}{UTF8}{min}見える\end{CJK}} & \textbf{\begin{CJK}{UTF8}{mj}보이는\end{CJK}} \\*
\textit{invisible} & \textit{invisible} & \textit{unsichtbar} & \textit{invisible} & \textit{\begin{CJK}{UTF8}{gbsn}不可见\end{CJK}} & \textit{\begin{CJK}{UTF8}{min}見えない\end{CJK}} & \textit{\begin{CJK}{UTF8}{mj}보이지 않는\end{CJK}} \\ \midrule
\textbf{warm} & \textbf{chaud} & \textbf{warm} & \textbf{cálido} & \textbf{\begin{CJK}{UTF8}{gbsn}暖和\end{CJK}} & \textbf{\begin{CJK}{UTF8}{min}暖かい\end{CJK}} & \textbf{\begin{CJK}{UTF8}{mj}따뜻한\end{CJK}} \\*
\textit{cool, cold} & \textit{frais, froid} & \textit{kühl, kalt} & \textit{fresco, frío} & \textit{\begin{CJK}{UTF8}{gbsn}凉, 冷\end{CJK}} & \textit{\begin{CJK}{UTF8}{min}涼しい, 冷たい\end{CJK}} & \textit{\begin{CJK}{UTF8}{mj}시원한, 차가운\end{CJK}} \\ \midrule
\textbf{smooth} & \textbf{lisse} & \textbf{glatt} & \textbf{liso} & \textbf{\begin{CJK}{UTF8}{gbsn}光滑\end{CJK}} & \textbf{\begin{CJK}{UTF8}{min}滑らかな\end{CJK}} & \textbf{\begin{CJK}{UTF8}{mj}매끄러운\end{CJK}} \\*
\textit{rough, bumpy} & \textit{rugueux, bosselé} & \textit{rau, holprig} & \textit{áspero, irregular} & \textit{\begin{CJK}{UTF8}{gbsn}粗糙\end{CJK}} & \textit{\begin{CJK}{UTF8}{min}粗い\end{CJK}} & \textit{\begin{CJK}{UTF8}{mj}거친, 울퉁불퉁한\end{CJK}} \\ \midrule
\textbf{thin} & \textbf{mince} & \textbf{dünn} & \textbf{delgado} & \textbf{\begin{CJK}{UTF8}{gbsn}薄\end{CJK}} & \textbf{\begin{CJK}{UTF8}{min}薄い\end{CJK}} & \textbf{\begin{CJK}{UTF8}{mj}얇은\end{CJK}} \\*
\textit{thick, fat} & \textit{épais} & \textit{dick} & \textit{grueso} & \textit{\begin{CJK}{UTF8}{gbsn}厚\end{CJK}} & \textit{\begin{CJK}{UTF8}{min}厚い\end{CJK}} & \textit{\begin{CJK}{UTF8}{mj}두꺼운\end{CJK}} \\ \midrule
\textbf{tall} & \textbf{grand} & \textbf{groß} & \textbf{alto} & \textbf{\begin{CJK}{UTF8}{gbsn}高\end{CJK}} & \textbf{\begin{CJK}{UTF8}{min}背が高い\end{CJK}} & \textbf{\begin{CJK}{UTF8}{mj}키가 큰\end{CJK}} \\*
\textit{short} & \textit{petit} & \textit{klein} & \textit{bajo} & \textit{\begin{CJK}{UTF8}{gbsn}矮\end{CJK}} & \textit{\begin{CJK}{UTF8}{min}背が低い\end{CJK}} & \textit{\begin{CJK}{UTF8}{mj}키가 작은\end{CJK}} \\ \midrule
\textbf{married} & \textbf{marié} & \textbf{verheiratet} & \textbf{casado} & \textbf{\begin{CJK}{UTF8}{gbsn}已婚\end{CJK}} & \textbf{\begin{CJK}{UTF8}{min}既婚の\end{CJK}} & \textbf{\begin{CJK}{UTF8}{mj}결혼한\end{CJK}} \\*
\textit{single, unmarried} & \textit{célibataire} & \textit{ledig, unverheiratet} & \textit{soltero} & \textit{\begin{CJK}{UTF8}{gbsn}单身, 未婚\end{CJK}} & \textit{\begin{CJK}{UTF8}{min}独身の, 未婚の\end{CJK}} & \textit{\begin{CJK}{UTF8}{mj}독신의, 미혼의\end{CJK}} \\ \midrule
\textbf{optimistic} & \textbf{optimiste} & \textbf{optimistisch} & \textbf{optimista} & \textbf{\begin{CJK}{UTF8}{gbsn}乐观\end{CJK}} & \textbf{\begin{CJK}{UTF8}{min}楽観的な\end{CJK}} & \textbf{\begin{CJK}{UTF8}{mj}낙관적인\end{CJK}} \\*
\textit{pessimistic} & \textit{pessimiste} & \textit{pessimistisch} & \textit{pesimista} & \textit{\begin{CJK}{UTF8}{gbsn}悲观\end{CJK}} & \textit{\begin{CJK}{UTF8}{min}悲観的な\end{CJK}} & \textit{\begin{CJK}{UTF8}{mj}비관적인\end{CJK}} \\ \midrule
\textbf{permanent} & \textbf{permanent} & \textbf{permanent} & \textbf{permanente} & \textbf{\begin{CJK}{UTF8}{gbsn}永久\end{CJK}} & \textbf{\begin{CJK}{UTF8}{min}恒久的な\end{CJK}} & \textbf{\begin{CJK}{UTF8}{mj}영구적인\end{CJK}} \\*
\textit{temporary} & \textit{temporaire} & \textit{temporär} & \textit{temporal} & \textit{\begin{CJK}{UTF8}{gbsn}临时\end{CJK}} & \textit{\begin{CJK}{UTF8}{min}一時的な\end{CJK}} & \textit{\begin{CJK}{UTF8}{mj}일시적인\end{CJK}} \\ \midrule
\textbf{present} & \textbf{présent} & \textbf{gegenwärtig} & \textbf{presente} & \textbf{\begin{CJK}{UTF8}{gbsn}现在\end{CJK}} & \textbf{\begin{CJK}{UTF8}{min}現在の\end{CJK}} & \textbf{\begin{CJK}{UTF8}{mj}현재의\end{CJK}} \\*
\textit{absent, past} & \textit{absent, passé} & \textit{abwesend, vergangen} & \textit{ausente, pasado} & \textit{\begin{CJK}{UTF8}{gbsn}缺席, 过去\end{CJK}} & \textit{\begin{CJK}{UTF8}{min}不在の, 過去の\end{CJK}} & \textit{\begin{CJK}{UTF8}{mj}부재의, 과거의\end{CJK}} \\ \midrule
\textbf{public} & \textbf{public} & \textbf{öffentlich} & \textbf{público} & \textbf{\begin{CJK}{UTF8}{gbsn}公共\end{CJK}} & \textbf{\begin{CJK}{UTF8}{min}公共の\end{CJK}} & \textbf{\begin{CJK}{UTF8}{mj}공공의\end{CJK}} \\*
\textit{private} & \textit{privé} & \textit{privat} & \textit{privado} & \textit{\begin{CJK}{UTF8}{gbsn}私人\end{CJK}} & \textit{\begin{CJK}{UTF8}{min}個人の\end{CJK}} & \textit{\begin{CJK}{UTF8}{mj}사적인\end{CJK}} \\ \midrule
\textbf{real} & \textbf{réel} & \textbf{echt} & \textbf{real} & \textbf{\begin{CJK}{UTF8}{gbsn}真实\end{CJK}} & \textbf{\begin{CJK}{UTF8}{min}本物の\end{CJK}} & \textbf{\begin{CJK}{UTF8}{mj}실제적인\end{CJK}} \\*
\textit{fake, unreal} & \textit{faux, irréel} & \textit{falsch, unreal} & \textit{falso, irreal} & \textit{\begin{CJK}{UTF8}{gbsn}假, 虚假\end{CJK}} & \textit{\begin{CJK}{UTF8}{min}偽物の, 非現実の\end{CJK}} & \textit{\begin{CJK}{UTF8}{mj}가짜의, 비현실적인\end{CJK}} \\ \midrule
\textbf{responsible} & \textbf{responsable} & \textbf{verantwortlich} & \textbf{responsable} & \textbf{\begin{CJK}{UTF8}{gbsn}负责\end{CJK}} & \textbf{\begin{CJK}{UTF8}{min}責任\end{CJK}} & \textbf{\begin{CJK}{UTF8}{mj}책임 있는\end{CJK}} \\*
\textit{irresponsible} & \textit{irresponsable} & \textit{unverantwortlich} & \textit{irresponsable} & \textit{\begin{CJK}{UTF8}{gbsn}不负责\end{CJK}} & \textit{\begin{CJK}{UTF8}{min}無責任な, 無責任\end{CJK}} & \textit{\begin{CJK}{UTF8}{mj}무책임한\end{CJK}} \\ \midrule
\textbf{single} & \textbf{célibataire} & \textbf{ledig} & \textbf{soltero} & \textbf{\begin{CJK}{UTF8}{gbsn}单身\end{CJK}} & \textbf{\begin{CJK}{UTF8}{min}独身の\end{CJK}} & \textbf{\begin{CJK}{UTF8}{mj}독신의\end{CJK}} \\*
\textit{married} & \textit{marié} & \textit{verheiratet} & \textit{casado} & \textit{\begin{CJK}{UTF8}{gbsn}已婚\end{CJK}} & \textit{\begin{CJK}{UTF8}{min}既婚の, 結婚している\end{CJK}} & \textit{\begin{CJK}{UTF8}{mj}결혼한\end{CJK}} \\ \midrule
\textbf{sour} & \textbf{acide} & \textbf{sauer} & \textbf{agrio} & \textbf{\begin{CJK}{UTF8}{gbsn}酸\end{CJK}} & \textbf{\begin{CJK}{UTF8}{min}酸っぱい\end{CJK}} & \textbf{\begin{CJK}{UTF8}{mj}신\end{CJK}} \\*
\textit{sweet} & \textit{doux} & \textit{süß} & \textit{dulce} & \textit{\begin{CJK}{UTF8}{gbsn}甜\end{CJK}} & \textit{\begin{CJK}{UTF8}{min}甘い\end{CJK}} & \textit{\begin{CJK}{UTF8}{mj}달콤한\end{CJK}} \\ \midrule
\textbf{useful} & \textbf{utile} & \textbf{nützlich} & \textbf{útil} & \textbf{\begin{CJK}{UTF8}{gbsn}有用\end{CJK}} & \textbf{\begin{CJK}{UTF8}{min}役に立つ\end{CJK}} & \textbf{\begin{CJK}{UTF8}{mj}유용한\end{CJK}} \\*
\textit{useless} & \textit{inutile} & \textit{nutzlos} & \textit{inútil} & \textit{\begin{CJK}{UTF8}{gbsn}没用\end{CJK}} & \textit{\begin{CJK}{UTF8}{min}役に立たない\end{CJK}} & \textit{\begin{CJK}{UTF8}{mj}쓸모없는\end{CJK}} \\ \midrule
\textbf{vertical} & \textbf{vertical} & \textbf{vertikal} & \textbf{vertical} & \textbf{\begin{CJK}{UTF8}{gbsn}垂直\end{CJK}} & \textbf{\begin{CJK}{UTF8}{min}垂直な\end{CJK}} & \textbf{\begin{CJK}{UTF8}{mj}수직의\end{CJK}} \\*
\textit{horizontal} & \textit{horizontal} & \textit{horizontal} & \textit{horizontal} & \textit{\begin{CJK}{UTF8}{gbsn}水平\end{CJK}} & \textit{\begin{CJK}{UTF8}{min}水平な\end{CJK}} & \textit{\begin{CJK}{UTF8}{mj}수평의\end{CJK}} \\ \midrule
\textbf{well} & \textbf{bien} & \textbf{gut} & \textbf{bien} & \textbf{\begin{CJK}{UTF8}{gbsn}好\end{CJK}} & \textbf{\begin{CJK}{UTF8}{min}良い\end{CJK}} & \textbf{\begin{CJK}{UTF8}{mj}잘\end{CJK}} \\*
\textit{poorly} & \textit{mal} & \textit{schlecht} & \textit{mal, pobremente} & \textit{\begin{CJK}{UTF8}{gbsn}差\end{CJK}} & \textit{\begin{CJK}{UTF8}{min}悪い\end{CJK}} & \textit{\begin{CJK}{UTF8}{mj}못, 형편없이\end{CJK}} \\ \midrule
\textbf{winning} & \textbf{gagnant} & \textbf{gewinnend} & \textbf{ganador} & \textbf{\begin{CJK}{UTF8}{gbsn}获胜\end{CJK}} & \textbf{\begin{CJK}{UTF8}{min}勝利の\end{CJK}} & \textbf{\begin{CJK}{UTF8}{mj}승리하는\end{CJK}} \\*
\textit{losing} & \textit{perdant} & \textit{verlierend} & \textit{perdedor} & \textit{\begin{CJK}{UTF8}{gbsn}失败\end{CJK}} & \textit{\begin{CJK}{UTF8}{min}敗北の\end{CJK}} & \textit{\begin{CJK}{UTF8}{mj}패배하는\end{CJK}} \\ \midrule
\textbf{unkind} & \textbf{méchant} & \textbf{unfreundlich} & \textbf{desagradable} & \textbf{\begin{CJK}{UTF8}{gbsn}不善良\end{CJK}} & \textbf{\begin{CJK}{UTF8}{min}意地悪な\end{CJK}} & \textbf{\begin{CJK}{UTF8}{mj}불친절한\end{CJK}} \\*
\textit{kind} & \textit{gentil} & \textit{freundlich} & \textit{amable} & \textit{\begin{CJK}{UTF8}{gbsn}善良\end{CJK}} & \textit{\begin{CJK}{UTF8}{min}親切な\end{CJK}} & \textit{\begin{CJK}{UTF8}{mj}친절한\end{CJK}} \\ \midrule
\textbf{cowardly} & \textbf{lâche} & \textbf{feige} & \textbf{cobarde} & \textbf{\begin{CJK}{UTF8}{gbsn}懦弱\end{CJK}} & \textbf{\begin{CJK}{UTF8}{min}臆病な\end{CJK}} & \textbf{\begin{CJK}{UTF8}{mj}겁많은\end{CJK}} \\*
\textit{brave} & \textit{courageux} & \textit{mutig} & \textit{valiente} & \textit{\begin{CJK}{UTF8}{gbsn}勇敢\end{CJK}} & \textit{\begin{CJK}{UTF8}{min}勇敢な\end{CJK}} & \textit{\begin{CJK}{UTF8}{mj}용감한\end{CJK}}
\end{longtable}
% }

\normalsize
\section{Listings Used in \textbf{Enumerations}}
\label{app:enumerations}
We prepare 7 ordered enumerations to use for the enumeration-structured sentences (\textit{The \{category\} are: \{choices\}}). The English dataset was created entirely by human annotators and then translated into the remaining languages following the same LLM-assisted translation and native-speaker validation process, using Gemini 3 Flash~\citep{deepmind2025gemini3flash} for translation. All the categories, listings, and the split of the listed choices and continuation words are shown in \Cref{tab:enumerations}.

% table* allows the table to span across both columns in a two-column document
\begin{table*}[t]
\centering
\caption{Enumerations Dataset: categories, items, and split thresholds.}
\label{tab:enumerations}
\scriptsize % Scriptsize helps fit the high volume of text across 7 languages
\begin{CJK*}{UTF8}{gbsn} % Defaulting to gbsn (Chinese) for the main environment
\begin{tabularx}{\textwidth}{l c l l X}
\toprule
\textbf{Category ID} & \textbf{Split} & \textbf{Lang} & \textbf{Category Name} & \textbf{Listings} \\ \midrule

% MONTHS
months & 4|8 & en & months of the year & January, February, March, April, May, June, July, August, September, October, November, December \\
 & & de & Monate des Jahres & Januar, Februar, März, April, Mai, Juni, Juli, August, September, Oktober, November, Dezember \\
 & & fr & mois de l'année & janvier, février, mars, avril, mai, juin, juillet, août, septembre, octobre, novembre, décembre \\
 & & es & meses del año & enero, febrero, marzo, abril, mayo, junio, julio, agosto, septiembre, octubre, noviembre, diciembre \\
 & & zh & 一年中的月份 & 一月, 二月, 三月, 四月, 五月, 六月, 七月, 八月, 九月, 十月, 十一月, 十二月 \\
 & & ja & \begin{CJK*}{UTF8}{min}一年の月\end{CJK*} & \begin{CJK*}{UTF8}{min}一月, 二月, 三月, 四月, 五月, 六月, 七月, 八月, 九月, 十月, 十一月, 十二月\end{CJK*} \\
 & & ko & \begin{CJK*}{UTF8}{mj}1년의 열두 달\end{CJK*} & \begin{CJK*}{UTF8}{mj}1월, 2월, 3월, 4월, 5월, 6월, 7월, 8월, 9월, 10월, 11월, 12월\end{CJK*} \\ \midrule

% NUMBERS
numbers & 4|6 & en & numbers 1 to 10 & one, two, three, four, five, six, seven, eight, nine, ten \\
 & & de & Zahlen von eins bis zehn & eins, zwei, drei, vier, fünf, sechs, sieben, acht, neun, zehn \\
 & & fr & nombres de un à dix & un, deux, trois, quatre, cinq, six, sept, huit, neuf, dix \\
 & & es & números del uno al diez & uno, dos, tres, cuatro, cinco, seis, siete, ocho, nueve, diez \\
 & & zh & 从一到十的数字 & 一, 二, 三, 四, 五, 六, 七, 八, 九, 十 \\
 & & ja & \begin{CJK*}{UTF8}{min}一から十までの数字\end{CJK*} & \begin{CJK*}{UTF8}{min}一, 二, 三, 四, 五, 六, 七, 八, 九, 十\end{CJK*} \\
 & & ko & \begin{CJK*}{UTF8}{mj}1부터 10까지의 숫자\end{CJK*} & \begin{CJK*}{UTF8}{mj}일, 이, 삼, 사, 오, 육, 칠, 팔, 구, 십\end{CJK*} \\ \midrule

% DAYS OF WEEK
days\_of\_week & 3|4 & en & days of the week & Sunday, Monday, Tuesday, Wednesday, Thursday, Friday, Saturday \\
 & & de & Wochentage & Sonntag, Montag, Dienstag, Mittwoch, Donnerstag, Freitag, Samstag \\
 & & fr & jours de la semaine & dimanche, lundi, mardi, mercredi, jeudi, vendredi, samedi \\
 & & es & días de la semana & domingo, lunes, martes, miércoles, jueves, viernes, sábado \\
 & & zh & 一星期中的日子 & 星期日, 星期一, 星期二, 星期三, 星期四, 星期五, 星期六 \\
 & & ja & \begin{CJK*}{UTF8}{min}曜日\end{CJK*} & \begin{CJK*}{UTF8}{min}日曜日, 月曜日, 火曜日, 水曜日, 木曜日, 金曜日, 土曜日\end{CJK*} \\
 & & ko & \begin{CJK*}{UTF8}{mj}요일\end{CJK*} & \begin{CJK*}{UTF8}{mj}일요일, 월요일, 화요일, 수요일, 목요일, 금요일, 토요일\end{CJK*} \\ \midrule

% FOUR SEASONS
four\_seasons & 2|2 & en & four seasons & spring, summer, fall, winter \\
 & & de & vier Jahreszeiten & Frühling, Sommer, Herbst, Winter \\
 & & fr & quatre saisons & printemps, été, automne, hiver \\
 & & es & cuatro estaciones & primavera, verano, otoño, invierno \\
 & & zh & 四季 & 春, 夏, 秋, 冬 \\
 & & ja & \begin{CJK*}{UTF8}{min}四季\end{CJK*} & \begin{CJK*}{UTF8}{min}春, 夏, 秋, 冬\end{CJK*} \\
 & & ko & \begin{CJK*}{UTF8}{mj}사계절\end{CJK*} & \begin{CJK*}{UTF8}{mj}봄, 여름, 가을, 겨울\end{CJK*} \\ \midrule

% TIMES OF DAY
times\_of\_day & 2|2 & en & times of day & morning, afternoon, evening, night \\
 & & de & Tageszeiten & Morgen, Nachmittag, Abend, Nacht \\
 & & fr & moments de la journée & matin, après-midi, soir, nuit \\
 & & es & momentos del día & mañana, tarde, tarde, noche \\
 & & zh & 一天的时段 & 早上, 下午, 晚上, 夜里 \\
 & & ja & \begin{CJK*}{UTF8}{min}一日の時間帯\end{CJK*} & \begin{CJK*}{UTF8}{min}朝, 午後, 晩, 夜\end{CJK*} \\
 & & ko & \begin{CJK*}{UTF8}{mj}하루의 시간대\end{CJK*} & \begin{CJK*}{UTF8}{mj}아침, 오후, 저녁, 밤\end{CJK*} \\ \midrule

% CARDINAL DIRECTIONS
cardinal\_directions & 2|2 & en & cardinal directions & North, South, East, West \\
 & & de & Himmelsrichtungen & Norden, Süden, Osten, Westen \\
 & & fr & points cardinaux & nord, sud, est, ouest \\
 & & es & puntos cardinales & norte, sur, este, oeste \\
 & & zh & 基本方位 & 北, 南, 东, 西 \\
 & & ja & \begin{CJK*}{UTF8}{min}基本方位\end{CJK*} & \begin{CJK*}{UTF8}{min}北, 南, 東, 西\end{CJK*} \\
 & & ko & \begin{CJK*}{UTF8}{mj}방위\end{CJK*} & \begin{CJK*}{UTF8}{mj}북, 남, 동, 서\end{CJK*} \\ \midrule

% COLORS
primary\_colors & 1|2 & en & primary colors of light & red, green, blue \\
 & & de & Primärfarben des Lichts & Rot, Grün, Blau \\
 & & fr & couleurs primaires de la lumière & rouge, vert, bleu \\
 & & es & colores primarios de la luz & rojo, verde, azul \\
 & & zh & 光的三原色 & 红, 绿, 蓝 \\
 & & ja & \begin{CJK*}{UTF8}{min}光の三原色\end{CJK*} & \begin{CJK*}{UTF8}{min}赤, 緑, 青\end{CJK*} \\
 & & ko & \begin{CJK*}{UTF8}{mj}빛의 삼원색\end{CJK*} & \begin{CJK*}{UTF8}{mj}빨강, 초록, 파랑\end{CJK*} \\ 
\bottomrule

\end{tabularx}
\end{CJK*}
\end{table*}

\normalsize
\section{Additional Redundancy Analysis}
\label{apx:redundancy}
To assess the equivalence of the latent sets identified by the three selection methods, we compute representative per-layer activation space directions that correspond to the decoder directions of these layer-wise latents, then calculate the pairwise cosine similarity between per-method directions. If the latents of two selection methods for the same language are identical or equivalent in their representation in a certain layer, we expect their corresponding residual stream directions to be similar (large cosine similarity), while a small cosine similarity in a certain layer might indicate divergence. We report in Table~\ref{tab:cosine-sim} for each pair of selection methods the minimum cosine similarity, as well as the across-layer average, for the language latents of \texttt{Gemma-2-2B} and \texttt{Qwen3-4B}.

\begin{table}[!htp]\centering\small
\caption{Minimum and average cosine similarity between representative layer-wise activation space directions per language for \texttt{Gemma-2-2B} (top) and \texttt{Qwen3-4B} (bottom).}
\resizebox{0.5\textwidth}{!}{ % use this if the table is too large
\begin{tabular}{c|l|r|ccccccc}\toprule
& &&es &en &zh &de &ja &fr &ko \\\midrule
\multirow{3}{.4cm}{\rotatebox[origin=c]{90}{\textbf{Gemma-2-2B~}}}&\texttt{AnnSel - ValSel} &Avg &0.66 &0.06 &0.58 &0.59 &0.71 &0.45 &0.30 \\
&&Min &0.47 &-0.1 &0.44 &0.41 &0.53 &0.07 &-0.07 \\ \cmidrule{2-10}
&\texttt{FreqSel - AnnSel} &Avg &0.57 &0.04 &0.59 &0.63 &0.47 &0.57 &0.41 \\
&&Min &0.09 &-0.15 &0.25 &0.24 &0.18 &0.12 &0.05 \\ \cmidrule{2-10}
&\texttt{ValSel - FreqSel} &Avg &0.34 &0.04 &0.45 &0.29 &0.42 &0.25 &0.30 \\
&&Min &-0.1 &-0.37 &0.11 &0.07 &0.08 &-0.05 &-0.44 \\ \cmidrule{2-10}
\specialrule{1.5pt}{0pt}{0pt} \\
\multirow{3}{.4cm}{\rotatebox[origin=c]{90}{\textbf{Qwen3-4B~~}}}&\texttt{AnnSel - ValSel} &Avg &0.07 &0.26 &0.47 &0.34 &0.35 &0.3 &0.62 \\ 
&&Min &-0.07 &-0.22 &0.06 &-0.04 &0.02 &-0.04 &0.31 \\ \cmidrule{2-10}
&\texttt{FreqSel - AnnSel} &Avg &0.41 &0.37 &0.38 &0.48 &0.55 &0.32 &0.35 \\
&&Min &-0.02 &0.04 &0.02 &0.23 &0.11 &0.00 &0.06 \\ \cmidrule{2-10}
&\texttt{ValSel - FreqSel} &Avg &0.42 &0.14 &0.21 &0.27 &0.46 &0.36 &0.35 \\
&&Min &-0.03 &-0.17 &-0.32 &-0.01 &0.00&-0.15 &-0.35 \\
\bottomrule
\end{tabular}
}
\label{tab:cosine-sim}
\end{table}

% \textcolor{red}{The generally low average cosine similarities (mostly below 0.4) and frequently negative minimum values confirm that the different methods are identifying latent subsets that correspond to genuinely distinct directions in activation space, not merely the same directions labelled differently. This suggests that language representations might exhibit multi-path redundancy in the latent space.}

Finally, we analyze the Neuronpedia annotations of the identified latent sets as a (potentially noisy) means to discern latent set overlap. We compute as a proxy the percentage of language latents whose Neuronpedia annotation refers explicitly to the language. While this is 100\% for \texttt{AnnSel} by design, the results for \texttt{ValSel} and \texttt{FreqSel} in \Cref{tab:language_annotation} reveal substantial variability.

\begin{table}[ht]
    \centering
        \caption{The percentage (\%) of extracted latents of \texttt{Gemma-2-2B} (top) and \texttt{Qwen3-4B} (bottom) per method and language whose Neuronpedia annotations explicitly refer to the language. \texttt{AnnSel} is 100\% by design.}
    \resizebox{0.5\textwidth}{!}{ 
    \begin{tabular}{c|r|ccccccc}
    \toprule
        &&en&fr&de&es& zh&ko&ja \\
        \midrule
       {\textbf{Gemma-2-2B~}}&\texttt{ValSel}  & 2 & 72 & 94 & 54 & 58 & 2 & 46 \\
        &\texttt{FreqSel} & 0.1 & 71 & 77 & 53 & 39 & 1 & 58 \\
        &\texttt{AnnSel} & \textit{100} & \textit{100}&\textit{100}&\textit{100}&\textit{100}&\textit{100}&\textit{100}\\
        \midrule
        {\textbf{Qwen3-4B~~}}&\texttt{ValSel}  & 2 &24 &32 &24 &12 &0 &8 \\
        &\texttt{FreqSel} & 1.1 &32.9 &38.2 &28.4 &37.5 &4.1 &16.7 \\
        &\texttt{AnnSel} & \textit{100} & \textit{100}&\textit{100}&\textit{100}&\textit{100}&\textit{100}&\textit{100}\\
        \bottomrule
    \end{tabular}
    }
    \label{tab:language_annotation}
\end{table}

\end{document}